\documentclass{dsme}

\usepackage[authoryear,sort&compress,round]{natbib}
\usepackage{amsmath,mathtools}
\usepackage{graphicx}
\usepackage{wrapfig}
\usepackage{placeins}
\usepackage{flushend}
\usepackage{siunitx}
\usepackage{soul}
\usepackage{todonotes}
\usepackage{comment}
\usepackage{algorithm}
\usepackage{algpseudocode}
\usepackage{pifont}
\usepackage{xspace}

\usetikzlibrary{arrows.meta,positioning,fit,calc,shapes.geometric,arrows,backgrounds}
\tcbuselibrary{breakable}

\setkeys{Gin}{width=\linewidth,keepaspectratio}

\renewcommand{\titlefont}{\color{black}\bfseries\fontsize{18}{21}\selectfont}
\renewcommand{\sectionfont}{\color{black}\bfseries\fontsize{12}{14}\selectfont}
\renewcommand{\subsectionfont}{\color{black}\bfseries\fontsize{10.5}{12.5}\selectfont}
\AtBeginDocument{\setlength{\bibsep}{0pt}}

\renewcommand{\cite}{\citep}

\makeatletter
\def\plist@algorithm{Alg.\space}
\makeatother

\renewcommand{\thepart}{\Alph{part}}
\titleformat{\part}[block]
  {\centering\color{black}\large\bfseries\scshape}
  {}
  {0pt}
  {\titlerule[0.8pt]\vspace{0.55ex}\partname~\thepart:\enspace#1\\[0.45ex]\titlerule[0.8pt]}
\titlespacing*{\part}{0pt}{1.6em plus 0.4em minus 0.2em}{0.9em}

\newcommand{\displayheadingformat}{\raggedright\hyphenpenalty=10000\exhyphenpenalty=10000}
\titleformat{\section}
  {\sectionfont\displayheadingformat}
  {\thesection}
  {1em}
  {#1}
\titleformat{\subsection}
  {\subsectionfont\displayheadingformat}
  {\thesubsection}
  {1em}
  {#1}
\titlespacing*{\section}{0pt}{2.2ex plus 0.6ex minus 0.2ex}{0.9ex plus 0.2ex}
\titlespacing*{\subsection}{0pt}{1.8ex plus 0.5ex minus 0.2ex}{0.65ex plus 0.15ex}

\definecolor{RWTHBlue}{rgb}{0, 0.32941176470588235, 0.6235294117647059}
\definecolor{RWTHBordeaux}{rgb}{0.6313725490196078, 0.06274509803921569, 0.20784313725490197}
\definecolor{RWTHRed75}{rgb}{0.84705882352941175,0.36078431372549019,0.25490196078431371}
\definecolor{RWTHRed}{rgb}{0.8,0.027450980392156862,0.11764705882352941}
\definecolor{RWTHRed50}{rgb}{0.90196078431372551,0.58823529411764708,0.47450980392156861}
\definecolor{RWTHRed25}{rgb}{0.95294117647058818,0.803921568627451,0.73333333333333328}
\definecolor{RWTHOrange}{rgb}{0.9647058823529412, 0.6588235294117647, 0}
\definecolor{RWTHGreen}{rgb}{0.3411764705882353, 0.6705882352941176, 0.15294117647058825}
\definecolor{RWTHViolett}{rgb}{0.38039215686274508, 0.12941176470588237, 0.34509803921568627}
\definecolor{RWTHTuerkis}{rgb}{0.0, 0.596078431372549, 0.63137254901960782}
\definecolor{RWTHPetrol}{rgb}{0.0, 0.38039215686274508, 0.396078431372549}

\definecolor{linkblue}{rgb}{0.008, 0.4, 0.733}

\newcommand{\ie}{i\/.\/e\/.,\/~}
\newcommand{\eg}{e\/.\/g\/.,\/~}
\newcommand{\cf}{cf\/.\/~}

\newcommand{\argmin}[1]{\mathrm{arg}\min_{#1}} %

\newcommand*{\obj}{\ensuremath{J}} %

\newcommand{\boiter}{\ensuremath{k}}
\newcommand{\maxboiter}{\ensuremath{K}}
\newcommand{\obsnoise}{\ensuremath{\epsilon}}
\newcommand{\bodata}{\ensuremath{\mathcal{D}}}
\newcommand{\gphyp}{\ensuremath{\lambda_\mathrm{GPR}}}

\newcommand{\params}{\boldsymbol{\theta}} %
\newcommand{\paramSet}{\ensuremath{\Theta}} %

\newcommand*{\state}{\ensuremath{\boldsymbol{x}}}
\newcommand*{\statedomain}{\ensuremath\mathcal{X}}
\newcommand*{\ssoutput}{\ensuremath{\boldsymbol{y}}}
\newcommand*{\ssoutputdomain}{\ensuremath{\mathcal{Y}}}
\newcommand*{\act}{\ensuremath{\boldsymbol{u}}}
\newcommand*{\actdomain}{\ensuremath\mathcal{U}}
\newcommand*{\dist}{\ensuremath{\boldsymbol{w}}}
\newcommand*{\distdomain}{\ensuremath{\mathcal{W}}}
\newcommand*{\ts}{\ensuremath{t}}
\newcommand*{\trajectory}{\ensuremath{\tau_{\params}}}
\newcommand*{\trajectoryk}{\ensuremath{\tau_{\params_k}}}

\newcommand{\context}{\mathbf{s}}

\newcommand*{\policy}{\ensuremath{\pi}}

\newtcolorbox{examplebox}[1]{colback=blue!5!white,
colframe=blue!75!black,fonttitle=\bfseries,
title={#1},boxrule=0pt,breakable}

\newtcolorbox{recombox}[1]{colback=blue!5!white,
colframe=blue!75!black,fonttitle=\bfseries,
title={#1},boxrule=0pt,breakable}

\newtcolorbox{probbox}[1]{colback=blue!5!white,
colframe=blue!75!black,fonttitle=\bfseries,
title={#1},boxrule=0pt,breakable}

\usepackage{enumitem}
\usepackage{pifont}

\newcommand{\checkbox}{\ding{113}} %

\newlist{checklist}{itemize}{1}
\setlist[checklist]{
  label=\checkbox,
  leftmargin=2em,
  itemsep=2pt,
  topsep=2pt,
  parsep=0pt
}

\definecolor{myblue}{HTML}{D2E1F2}
\definecolor{mygreen}{HTML}{C5E7CA}
\definecolor{myorange}{HTML}{FFF0D5}
\definecolor{mygrey}{HTML}{E5E5E5}

\tcbset{
    lavenderbox/.style={
            colback=Lavender!30,
            colframe=Lavender!80!black!50,
            coltitle=black,
            fonttitle=\bfseries,
            boxrule=0pt,
            arc=5pt,
            left=5pt,
            right=5pt,
            top=-5pt,
              breakable,
            bottom=0pt
        },
    lavenderboxsmall/.style={
            colback=Lavender!30,
            colframe=Lavender!80!black!50,
            coltitle=black,
            fonttitle=\bfseries,
            boxrule=0pt,
            arc=5pt,
            left=5pt,
            right=5pt,
            top=0pt,
              breakable,
            bottom=0pt
        },
    bluebox/.style={
            colback=myblue!20,
            colbacktitle=myblue,
            colframe=myblue!70!black,
            coltitle=black,
            fonttitle=\bfseries,
            boxrule=0pt,
            arc=5pt,
            left=5pt,
            right=5pt,
            top=5pt,
              breakable,
            bottom=5pt
        },
    methodbox/.style={
            colback=myblue!14,
            colframe=myblue!65!black,
            coltitle=black,
            boxrule=0pt,
            leftrule=2.2pt,
            arc=5pt,
            left=7pt,
            right=7pt,
            top=5pt,
            bottom=5pt,
            before skip=0.1em,
            after skip=0.9em,
            breakable
        },
    greenbox/.style={
            colback=mygreen!20,
            colbacktitle=mygreen,
            colframe=mygreen!60!black,
            coltitle=black,
            fonttitle=\bfseries,
            boxrule=0pt,
            arc=5pt,
            left=5pt,
            right=5pt,
            top=5pt,
              breakable,
            bottom=5pt
        },
    orangebox/.style={
            colback=myorange!30,
            colbacktitle=myorange,
            colframe=myorange!80!black,
            coltitle=black,
            fonttitle=\bfseries,
            boxrule=0pt,
            arc=5pt,
            left=5pt,
            right=5pt,
            top=5pt,
              breakable,
            bottom=5pt
        },
    greybox/.style={
            colback=mygrey!25,
            colbacktitle=mygrey,
            colframe=mygrey!70!black,
            coltitle=black,
            fonttitle=\bfseries,
            boxrule=0pt,
            arc=5pt,
            left=5pt,
            right=5pt,
            top=5pt,
              breakable,
            bottom=5pt
        },
    yellowbox/.style={
            colback=Yellow!20,
            colframe=Yellow!50!black,
            coltitle=black,
            fonttitle=\bfseries,
            boxrule=0pt,
            arc=5pt,
            left=5pt,
            right=5pt,
            top=5pt,
              breakable,
            bottom=5pt
        }
}

\newcommand{\fakepar}[1]{\vspace{1mm}
\noindent\textbf{#1.}\hspace{0.5em}}

\definecolor{circlegray}{HTML}{C2C2C2}

\DeclareRobustCommand{\circstep}[1]{%
    \tikz[baseline=(char.base)]{
        \node[shape=circle, draw=black, fill=circlegray, inner sep=0.5pt, minimum size=8pt] (char) {\textbf{#1}};
    }%
}

\usepackage{ifthen}
\newboolean{authnotes}
\setboolean{authnotes}{true}
\ifthenelse{\boolean{authnotes}}
{
    \newcommand{\drafttext}[1]{{\color{white!60!black} {\bf Text:} #1}}
    \newcommand{\needref}[1]{\textcolor{red}{[REF: #1]}}
    \newcommand{\todop}[1]{\textcolor{red}{[TODO: #1]}}
}
{
    \newcommand{\drafttext}[1]{}
    \newcommand{\needref}[1]{}
    \newcommand{\todop}[1]{}
}

\title{A Decade of Bayesian Optimization for Controller Tuning and Robot
Learning: Tutorial, Review, and Future Prospects}

\author[1]{{David Stenger}}
\author[1]{{Paul Brunzema}}
\author[1]{{Johanna Menn}}
\author[2]{{Alexander {von Rohr}}}
\author[3]{{\mbox{Angela P. Schoellig}}}
\author[1]{{\mbox{Sebastian Trimpe}}}

\makeatletter
\@namedef{@sep5}{\authorcr}
\makeatother

\affil[1]{RWTH Aachen University, Aachen, Germany}
\affil[2]{University of Technology Nuremberg, Nuremberg, Germany}
\affil[3]{Technical University of Munich, Munich, Germany}

\correspondingauthor{David Stenger, \href{mailto:david.stenger@dsme.rwth-aachen.de}{david.stenger@dsme.rwth-aachen.de}}
\renewcommand{\shortauthors}{D. Stenger et~al.}
\hypersetup{
  pdftitle={A Decade of Bayesian Optimization for Controller Tuning and Robot Learning: Tutorial, Review, and Future Prospects},
  pdfauthor={David Stenger, Paul Brunzema, Johanna Menn, Alexander von Rohr, Angela P. Schoellig, Sebastian Trimpe}
}

\begin{document}

\twocolumn[
\maketitle
\begin{abstract}
In the past decade, Bayesian optimization (BO) has emerged as a powerful and adaptable framework for automatic controller tuning and robot learning. This article offers a comprehensive overview of the state-of-the-art in BO, designed to support both researchers and practitioners in understanding recent advancements, practical applications, and future research directions.
We begin by adopting a practitioner's perspective, illustrating how to effectively set up BO through a representative controller tuning example. We position BO within the broader context of learning paradigms, ranging from deep reinforcement learning to data-driven control, and highlight scenarios where BO is most advantageous.
Next, we discuss the diverse range of BO methods that have been developed to tackle complex problems and specific applications. This article provides a unified perspective on the current landscape of BO, emphasizing its relevance to control systems and robotics, and it highlights future prospects by identifying key research challenges and promising avenues for advancing BO in the field.
This includes addressing
a significant gap in the BO landscape: the lack of standardized benchmark problems specifically for control-related applications. To foster future research and ensure rigorous evaluation, we start an effort towards a lightweight benchmark suite for control engineering and robotics. We also present metrics and best practices to facilitate direct comparisons between new BO algorithms and established state-of-the-art methods.

\end{abstract}
]
\begin{NoHyper}
\begingroup
\renewcommand{\thefootnote}{}
\footnotetext{\textbf{Funding.} This work has in part been supported by the German Research Foundation (DFG) within grant TR 1433/4-1 and under Germany’s Excellence Strategy EXC-2023 Internet of Production (390621612), by the Helmholtz School for Data Science in Life, Earth and Energy (HDS-LEE), and by the Robotics Institute Germany (RIG), funded through BMFTR grant 16ME0997K.
}
\endgroup
\end{NoHyper}

\section{Introduction}%

Sequential experimental design, and specifically\linebreak Bayesian optimization (BO), has emerged as an important but often hidden method for efficient and automated experimentation in various scientific and industrial applications. For example, Google's black-box optimization service, Vizier \cite{golovin2017google}, which employs BO, has optimized parameters for more than 70 million problems \cite{song2014vizier}, illustrating the substantial but mostly unseen practical impact of this method. 

\begin{figure}[t]
    \centering
    \includegraphics[width=0.98\columnwidth]{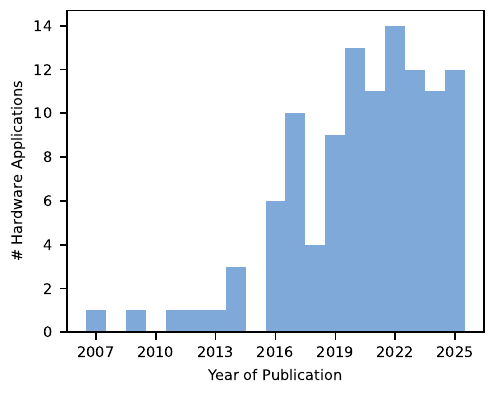}
    \caption{Number of control and robotics papers that use BO for online parameter tuning on hardware according to the review method described in Sec. \ref{sec:p2reviewmethod}. Total number of reviewed papers: 110.} %
    \label{fig:statistics_pub_year}
\end{figure}

In control and robotics, accurate parameter tuning is essential. Controllers ranging from classical PID structures to modern model predictive control (MPC) require careful tuning to achieve optimal performance. Traditionally, this tuning relies on domain expertise, trial-and-error experimentation, and heuristic methods, often consuming considerable time and resources. In the past decade, BO has become a widely recognized method for automating this tuning process, significantly reducing the need for manual intervention and offering a systematic approach to parameter optimization (see Fig. \ref{fig:statistics_pub_year}). 
Bayesian optimization distinguishes itself from other optimization methods mainly due to its data efficiency, particularly valuable when experiments are costly in terms of computation, time, or hardware usage. Unlike classical controller tuning methods, BO makes minimal assumptions about the system, treating it as a black-box requiring only measurable performance outputs. In contrast to deep reinforcement learning, BO leverages well established controller structures including their theoretical properties and thus requires little environment interactions. BO balances exploration of new parameter regions with exploitation of promising areas, guided by probabilistic models and information-theoretic criteria.

The versatility of the BO framework has led to the development of numerous methodological advancements addressing practical challenges specific to controller tuning and robotics. These advancements include methods for safe exploration to prevent hardware failures, handling constraints, multi-objective optimization scenarios, and adapting to changing environments.

Despite the widespread use and capabilities of BO, the lack of standardized benchmark problems and systematic evaluations in the control and robotics communities remains a challenge. Recognizing this gap, this paper provides an extensive survey of the current state-of-the-art and introduces evaluation guidelines and a step towards a lightweight benchmark suite for black-box controller tuning. These benchmarks enable fair and direct comparisons, encouraging further innovation and rigorous testing of new BO methods in controller tuning and robotic applications.

In this survey and tutorial, we provide guidance to both new and experienced practitioners in applying BO to controller tuning and robot learning. The paper starts with a practical tutorial illustrated by examples and progressively transitions to a comprehensive review of advanced BO methods and their applications. Finally, we highlight current research challenges and suggest promising directions for future work to further establish BO as a fundamental method for automated experimentation in control systems and robotics.

\subsection{Contributions}%

This article provides a comprehensive survey and tutorial on the state-of-the-art use of BO, specifically tailored for controller tuning and robot learning. We systematically cover both methods and practical considerations. Further, we provide benchmarking guidelines and outline emerging directions, fostering future BO research in control and robotics. Our contributions are structured into three main parts:

\fakepar{Part A: Practitioner-Oriented Tutorial}
We present a structured, practical tutorial demonstrating how typical controller tuning tasks can be formulated as black-box optimization problems and solved efficiently using BO. Through intuitive examples and accompanying code, readers can quickly understand core concepts and readily apply BO to their own control problems (Sec. \ref{sec:p1problemstatement} \& \ref{sec:p1BOandGPRIntro}). Additionally, we explicitly position BO relative to alternative learning paradigms, clarifying scenarios where BO offers advantages (Sec. \ref{sec:p1otherparadigms}).

\fakepar{Part B: Comprehensive Review of Methodologies and Applications}
We systematically review key design choices in BO, providing clear guidance on methodological best practices and their implementation (Sec.~\ref{sec:p2vanillaBO}). Advanced BO variants that address common real-world controller tuning challenges such as constraints, multiple objectives, crash constraints, and time-varying conditions are critically evaluated and contextualized within control applications (Sec.~\ref{sec:p2advanced_variants}). Furthermore, our extensive review of the literature covers 110 empirical hardware studies, synthesizing practical evidence and highlighting the real-world effectiveness and versatility of BO approaches (Sec.~\ref{sec:p2applications}).

\fakepar{Part C: Benchmarking Guidelines and Example}
Recognizing a critical gap in standardized evaluation, we provide best practices (Sec.~\ref{p3:howtobenchmark}) in benchmarking black-box optimization algorithms regarding baselines, metrics and statistical analysis. Additionally, we provide \textsc{TuneControl} (\href{https://github.com/Data-Science-in-Mechanical-Engineering/tunecontrol}{\faGithub}): a step towards a lightweight benchmark suite for black-box controller tuning. TuneControl is easily extensible, and we invite the community to contribute additional tuning tasks. We illustrate those best practices providing an example with code (Sec.~\ref{sec:p3examplebenchmark}).

To conclude, we summarize directions for future research and open issues in Sec.~\ref{sec:p3futureissues}. Compared to previous surveys, ours explicitly integrates theoretical clarity with actionable insights, advanced BO methodologies, and the development of dedicated benchmarks, uniquely tailored to the needs of the control and robotics communities.

\subsection{A Brief History of BO for Controller Tuning and Robot Learning} \label{sec:p0history} %

The foundations of BO have evolved significantly over the past decades. The earliest roots of modern BO trace back at least to 1964, when \citet{kushner1964new} introduced probabilistic improvement methods based on Wiener processes. A decade later, the widely-used expected improvement acquisition function was proposed by \citet{movckus1974bayesian}, while Gaussian process (GP) priors for BO emerged prominently by the late 1970s (\eg \cite{oohagan1978curve}). BO saw a significant increase in popularity following the seminal contributions of the design and analysis of computer experiments (DACE) framework by  \citet{sacks1989designs} and efficient global optimization (EGO) introduced by  \citet{jones1998efficient}. A comprehensive early survey summarizing Kriging-based methods was provided by \citet{jones2001taxonomy}. Readers interested in a detailed historical perspective are referred to \cite{garnett2022bayesian}.

The adoption of BO in control systems and robotics took additional time.
The earliest hardware application of BO in these domains was in 2007, with \citet{lizotte2007automatic} optimizing robot gait parameters and demonstrating clear advantages over classical hill-climbing methods. Following this breakthrough, BO received continued attention for gait optimization tasks (\eg \cite{hemker2009efficient,tesch2011using,tesch2013expensive}). A notable early tutorial, which among other topics addressed hierarchical reinforcement learning through BO, was provided by \citet{brochu2010tutorial}.

In 2013, BO researchers began exploring beyond traditional single-objective optimization, particularly by introducing multi-objective BO methods applied to robotic gait optimization \cite{tesch2013expensive}. A few years later, BO entered classical continuous control applications, first demonstrated with linear-quadratic regulator (LQR) tuning by \citet{marco2016automatic} and proportional-integral (PI) parameter optimization by \citet{berkenkamp2016safe}. The latter study notably introduced the influential paradigm of \emph{safe BO} to the control and robotics community.

Since these early demonstrations, BO’s popularity for controller tuning and robot learning has grown substantially, making it today an established and widely adopted method within control systems and robotics research.  

\subsection{Related Reviews and Tutorials} \label{sec:p0relatedwork}%

In addition to the recent book summarizing BO by Garnett~\cite{garnett2022bayesian}, numerous reviews and tutorials have been published on BO. Most existing surveys \cite{shahriari2016taking, frazier2018tutorial,candelieri2021gentle, bai2023transfer, greenhill2020bayesian, wang2022bayesian, astudilo2021thinking, wang2023recent} have a broad view and are not focused on control systems or robotics.  Consequently, these reviews neither systematically cover control-specific applications nor discuss the particular challenges encountered in real-world control systems.

Within the robotics and control domains specifically, comprehensive overviews remain scarce. After the early tutorial by \citet{brochu2010tutorial}, a recent tutorial by \citet{paulson2023tutorial} reviewed derivative-free policy learning with a particular emphasis on interpretable controller structures. In contrast, our paper explicitly targets a broader range of BO methodologies essential for robotics and control. Specifically, we clearly distinguish among advanced BO variants including constrained BO, safe BO, crash constraints, multi-objective BO, contextual, and time-varying BO each highly relevant but typically overlooked or conflated in previous reviews. Additionally, we provide benchmarking best practices to systematically evaluate and compare BO algorithms, addressing a significant practical gap that has not previously been covered.

\part{Tutorial}

Part A primarily addresses practitioners and researchers new to BO and explains when and how to use BO for controller tuning and robot learning. For that purpose, Sec.~\ref{sec:p1problemstatement} formulates the controller tuning problem as a black-box optimization problem using an illustrative example. Afterwards, Sec.~\ref{sec:p1BOandGPRIntro} introduces controller tuning with BO using the same illustrative example. Sec.~\ref{sec:p1problemstatement} and \ref{sec:p1BOandGPRIntro} consider the easiest form of BO: single objective BO with known feasible domain (``vanilla BO''). More advanced problem formulations are introduced in Sec.~\ref{sec:p2advanced_variants} of Part B. Finally, Sec.~\ref{sec:p1otherparadigms} distinguishes BO from other learning paradigms in control and provides recommendations on when to use BO.

\section{Controller Tuning as Black-Box Optimization} %

\label{sec:p1problemstatement}

\begin{figure*}[t]
    \centering
    \includegraphics[]{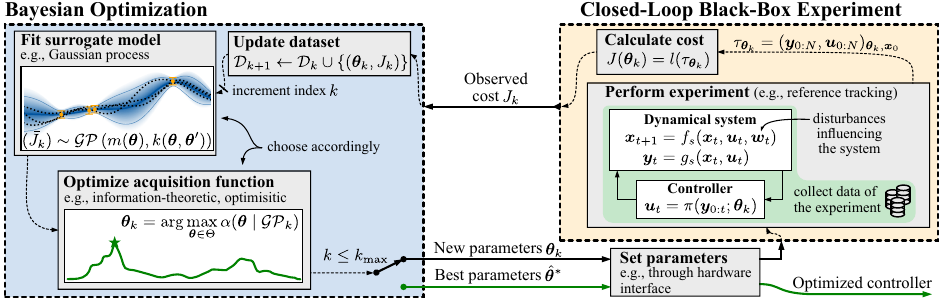}%
    \caption{Controller tuning with BO: The controller tuning task can be formulated as a black-box optimization problem (Sec.~\ref{sec:p1problemstatement}) in which the controller parameters are optimized with BO (Sec.~\ref{sec:p1BOandGPRIntro}).  
    }
    \label{fig:bo_for_control}
\end{figure*}

Sec.~\ref{sec:p1problemsingleobj} introduces the core idea of controller tuning as a black-box optimization problem.
The goal of this section is to introduce the mathematical notation and establish the black-box approach as a general tool to adjust closed-loop behaviors of arbitrary control systems using experimental data. Figure~\ref{fig:bo_for_control} summarizes the general idea. Sec.~\ref{sec:p1priorknowledge} gives an overview on how to extend this problem to more practical use cases and shows how to include prior knowledge.

\begin{tcolorbox}[greybox, title={Key Messages}]

The tuning objective is defined by some metric on the input/output trajectory allowing the optimization of arbitrary and unknown closed-loop systems (Sec.~\ref{sec:p1problemsingleobj}). The single-objective unconstrained problem formulation can be extended to more practical problem formulations and the Bayesian nature of BO allows for introducing prior knowledge in many ways (Sec.~\ref{sec:p1priorknowledge}).   

\end{tcolorbox}

\subsection{Single-Objective Unconstrained Problem Formulation} \label{sec:p1problemsingleobj}

To make controller tuning amenable to black-box optimization, we define the tuning task in terms of an objective function of a closed-loop trajectory. Crucially, all we assume about the control-loop is that we can measure its performance, and that this performance is influenced by a set of adjustable parameters. This generality allows controller tuning with BO for many relevant scenarios. For instance, we do not need to assume that the system is Markovian or that the controller is stateless or differentiable. We can also tune multiple controllers and observers at the same time, for example, when there are hierarchical control loops operating at different frequencies.
In fact, we do not even need to know what the control law is as long as we can choose some of its parameters and observe its effects in the closed-loop. This also allows tuning of proprietary controllers. We may even retain the theoretical guarantees of the optimized controllers.

The wide applicability of BO allows for great generality in its formalization. In the following we formalize the problem in a notation familiar for control engineers, but deviate slightly when another notation is helpful in the context of this tutorial.

We start with a discrete-time dynamical system 
\begin{equation} \label{eq:sec2systemdynamics}
\begin{aligned}
	\state_{\ts+1} &= f_{s}(\state_{\ts}, \act_{\ts}, \dist_\ts), \\
    \ssoutput_{\ts}  &= g_{s}(\state_{\ts},\act_{\ts}),
    \end{aligned}
\end{equation}
where $\state_\ts \in \statedomain$ is the state of the system, $\ssoutput_\ts \in \ssoutputdomain$ its measured output, $\act_\ts \in  \actdomain$ its input, $\dist_\ts \in \distdomain$ a disturbance, and $\state_0$ is the (possibly random) initial state at time step $\ts \in \mathbb{N}$. Here, $\statedomain$, $\actdomain$, and $\distdomain$ denote the state, input, and disturbance space, respectively. 
The system operates in closed loop with a controller $\policy$ determining the next input
	$\act_\ts = \policy(\ssoutput_{0:\ts}; \params)$
based on past measured outputs, where $\params \in \Theta$ are the parameters of the controller to be tuned and $\paramSet$ is the set of possible parameters, the search domain. This formulation includes the special cases of acting on the last output measurement $\pi(\ssoutput_t; \theta)$, state feedback $\pi(\ssoutput_t; \theta)$ with $\ssoutput_t = \state_t$, and stateful controllers. The parameters $\params$ may represent the gains of a PID,  poles of controller and observer, parameters of an LQR cost, and any other controller parameter (see Sec. \ref{sec:p2reviewcontrollers}).
In practice, control engineering expertise is often used to define a suitable parameterization and search domain. 

The tuning task is defined in terms of input/output trajectories generated by the closed-loop system, denoted as 
\begin{equation} \label{eq:sec2episode}
	\trajectory = (\ssoutput_{0:N}, \act_{0:N})_{\params,\state_0},
\end{equation}
where $N$ is the length of an episode. This implies that we do not need to know $f_s,g_s,\state_0$ or the state trajectory. We write $\trajectory$ to denote that the trajectory depends on the parameterization of the controller. The trajectory might be a random variable if initial state, disturbances, output measurement, or the controller are stochastic.
The function $l(\trajectory)$ takes a trajectory as input and calculates a cost metric.
Because the trajectory depends on the controller parameters $\params$, we can formulate the objective  directly as an unknown function $J(\params)$ of the parameters:
\begin{equation} \label{eq:sec2episodeshort}
	J(\params) = l(\trajectory),
\end{equation}
The tuning problem is to find optimal parameters $\params^*$ that minimize the objective function: 
\begin{align} \label{eq:single_objective}
    \params^* & =  \mathrm{argmin}_{\params \in \paramSet} \mathbb{E}\left[ \obj(\params) \right]. %
\end{align}

As a notational shorthand, we define $\bar{\obj}(\params) \coloneqq \mathbb{E}\left[ \obj(\params) \right]$ as the expected cost.

To approximately solve \eqref{eq:single_objective}, we assume that we can perform a closed-loop experiment for arbitrary controller parametrization $\params \in \paramSet$. The result of experiment $k$ with parameters $\params_k$ is a noisy observation of the tuning objective
    $\obj_\boiter = \bar{\obj}(\params_\boiter) + \obsnoise_\boiter$,
where $\obsnoise_\boiter$ is a random realization of, \eg random disturbances or initial state.
Given a trajectory we need the expression $l(\cdot)$ to calculate $J_k = l(\trajectoryk)$ but we may not be able to compute $J(\params)$ directly if, for example, the system is partially unknown. 
The resulting data set after $M$ evaluations used to inform the tuning process is $\mathcal{D} = \{(\params_\boiter, \obj_\boiter)\}^M_{k=1}$.
Further, we assume that experiments are costly for example in terms of time.

Consequently, we cast the tuning problem in \eqref{eq:single_objective} as a black-box optimization problem: We approximately solve \eqref{eq:single_objective} by sequentially performing closed-loop experiments.
Performance evaluations of the closed-loop system are treated as noisy, zeroth-order oracle and BO is the algorithm of choice to discover suitable controllers in a data efficient manner, \ie with as few experiments as possible.

\begin{figure*}
\centering
\includegraphics[]{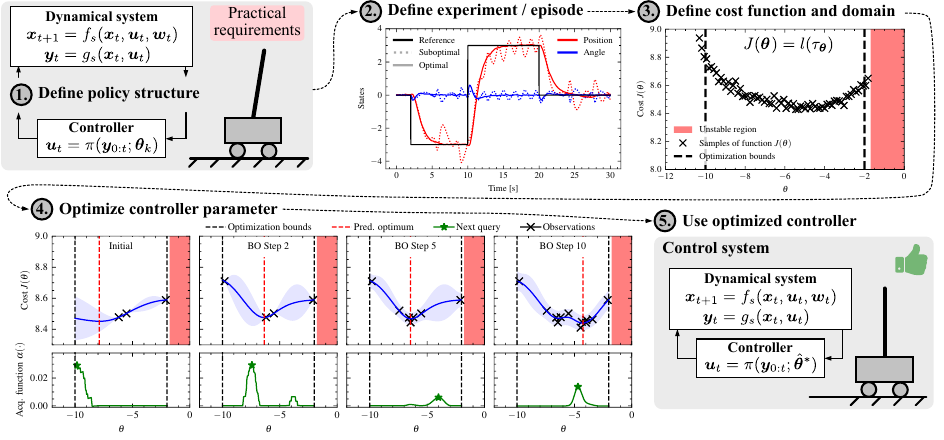}
    \caption{An example of BO for a cart pole system in 5 steps. Steps~{\scriptsize \circstep{1.}} to {\scriptsize \circstep{3.}} cover the problem formulation (Sec.~\ref{sec:p1problemstatement}) and Step~{\scriptsize \circstep{4.}} the optimization (Sec.~\ref{sec:p1BOandGPRIntro}). After optimization, the controller is operated with optimal parameters $\params^*$ in Step~{\scriptsize \circstep{5.}}.  \label{fig:bo_intuition}}   
\end{figure*}

\begin{tcolorbox}[bluebox, title={Illustrative cart-pole example (Fig. \ref{fig:bo_intuition}) \hfill {\it open in} \href{https://github.com/Data-Science-in-Mechanical-Engineering/tunecontrol/blob/main/examples/standard_bo_for_controller_tuning.ipynb}{\faGithub}}, label=examplebox1]

We consider of the well-known cart-pole as an illustrative example system (cf. Fig.\ref{fig:bo_intuition}).  
We deliberately do not report the system dynamics equations because they are irrelevant to setting up the black-box optimization problem. 
Informally, our task can be formulated as:
\begin{center}
    \begin{minipage}{0.95\textwidth}
    \emph{Stabilize the pendulum in the upright position while having the cart accurately track reference position jumps using a simple controller.}
    \end{minipage}
\end{center}
Below, we describe Steps {\footnotesize \circstep{1.}} to {\footnotesize \circstep{3.}} in Fig.~\ref{fig:bo_intuition}, which are necessary to formalize this problem as a black-box optimization problem, as illustrated in this section.
 
\medskip
\fakepar{\circstep{1.} Define policy structure}
We know from classical control engineering that a state feedback controller $u_k = -K \, \state_k$
can be used for this task. Our controller has 4 degrees of freedom $K =  \left[k_1, k_2, k_3, k_4 \right]$. For illustration purposes, we choose to only optimize one parameter $\params =  k_4 $. 

\medskip
\fakepar{\circstep{2.} Define experiment / episode}
The episode should be chosen in a way that closely resembles what we want to achieve with our system.
Here, we choose our reference trajectory $w_{x_1} (t)$ for the cart position to consist of reference jumps at three time steps.
The total episode takes 30 seconds and is depicted in Fig.~\ref{fig:bo_intuition}. 

\medskip
\fakepar{\circstep{3.} Define a cost function and domain}
We use a weighted version of the mean absolute error cost metric: 
$l(\tau_{\theta}) = \frac{1}{N} \sum_{k = 1}^{N} |\boldsymbol{q}^\top \state_k| + |R  u_k|$ 
where $\boldsymbol{q} = [10, 10, 10, 10]^\top$ and $R = 1$.
We bound the optimization domain to 
    $-10 \leq \theta \leq -2$.
As stated before, $l(\tau_{\theta})$ is a design choice and should be task-specific.

\smallskip
\begin{center}
    \textit{\dots example continues in Sec.~\ref{sec:p1BOIntro}\dots}
\end{center}

\end{tcolorbox}

\subsection{Real World Optimization Problems} \label{sec:p1priorknowledge}

Sec.~\ref{sec:p1problemsingleobj} considered the simplest problem formulation for controller tuning with BO: Single-objective unconstrained optimization without any prior knowledge. The BO framework offers many possibilities to extend this problem formulation to more advanced settings such as multiple objectives, constrained optimization, adaptive optimization or safe optimization. We discuss those extended formulations and methods to address them in Sec.~\ref{sec:p2advanced_variants}.   

Additionally, in control and robotics applications we often have prior knowledge about our system which we may want to use during the optimization process. This may either help to speed up optimization or make the optimization safer. BO offers many design choices and parameters that can be used to include different types of prior information. They are described in detail in the following sections. Fast-to-evaluate low-fidelity models of the system can be included in the probabilistic model of BO (see Sec.~\ref{sec:p2GPRmodel}). Slow-to-evaluate black-box models can be used in a multi-fidelity setting (see Sec.~\ref{sec:p2othervariants}). 
Subjective user beliefs can be leveraged through coactive optimization (see Sec.~\ref{sec:p2preferential}) 
or in the criterion to select new sample points (see Sec.~\ref{sec:p2acquisitionfun}).
Properties derived from analytical system models such as robust stability margins can be included via constraints (see Sec.~\ref{sec:p2Constrnt}) and any type of model can be used to select a starting parametrization for BO (see Sec.~\ref{sec:p2otherusecases}) and potentially combined with local (see Sec. \ref{sec:p2high_dim}) or safe (see Sec.~\ref{sec:p2safeBO}) optimization. We also refer to BO reviews that specifically consider transfer learning in BO \cite{bai2023transfer} and Gray-box BO \cite{astudilo2021thinking}, however not from a control and robotics perspective.

\section{Introduction to Bayesian Optimization and Gaussian Process Regression}%
\label{sec:p1BOandGPRIntro}

This section introduces the main concepts of BO with Gaussian process regression (GPR). We refer to \href{https://distill.pub/2020/bayesian-optimization/}{Exploring Bayesian Optimization} by \citet{agnihotri2020exploring} as an intuitive introductory resource. Sec.~\ref{sec:p1GPRIntro} introduces GPs as the most frequently used surrogate model in BO. Sec.~\ref{sec:p1BOIntro} introduces single-objective BO without any extensions. 
The most important degrees of freedom in BO for controller tuning and of the GP model are reviewed and discussed in detail in Sec. \ref{sec:p2vanillaBO}. Based on these discussions we give recommendations for the main degrees of freedom in Sec.~\ref{sec:p2vanillaBO}. For BO in more complex settings, we refer to \hbox{Sec.~\ref{sec:p2advanced_variants}}. 

\begin{tcolorbox}[greybox, title={Key Messages}]

Gaussian process regression is primarily used as a probabilistic non-parameteric surrogate model for the analytically unknown tuning objective (Sec.~\ref{sec:p1GPRIntro}). An acquisition function selects promising candidate parameters based on the GPR model. Those parameters are then evaluated and the GPR model is updated (Sec.~\ref{sec:p1BOIntro}).   

\end{tcolorbox}

\subsection{Gaussian Process Regression} \label{sec:p1GPRIntro}

Most work on BO for controller tuning uses GPR as a probabilistic, non-parametric surrogate for the unknown scalar objective $\obj(\params)$. Here we briefly summarize the GP model used throughout. For a comprehensive treatment, see \cite{rasmussen2006gaussian}.

\fakepar{Observation model}
In the noisy setting, we assume that an observation $\obj_i$ at parameter setting $\params_i$ is a corrupted measurement of an underlying latent function value $\bar{\obj}(\params_i)$,
\begin{equation} \label{eq:likelihood}
	\obj_i = \bar{\obj}(\params_i) + \obsnoise_i \mathrm{,}
\end{equation}
where the observation noise is commonly modeled as i.i.d.\ Gaussian, $\obsnoise_i \sim \mathcal{N}\!\left(0, {\sigma_\mathrm{n}}^2\right)$, yielding a homoscedastic Gaussian likelihood.

\fakepar{GP prior}
We place a GP prior on the latent objective function,
\begin{equation}
	\bar{\obj} \sim \mathcal{GP}\!\left(m, k\right),
\end{equation}
with mean function $m : \paramSet \rightarrow \mathbb{R}$ and covariance (kernel) function $k : \paramSet \times \paramSet \rightarrow \mathbb{R}$.
Equivalently, for any finite set of inputs, the corresponding latent function values are jointly Gaussian distributed.
In BO, this prior encodes structural assumptions about $\bar{\obj}$: $m$ specifies the prior expected objective value, and $k(\params_i,\params_j)$ specifies the prior covariance between $\bar{\obj}(\params_i)$ and $\bar{\obj}(\params_j)$.
For typical stationary kernels, $k(\params_i,\params_j)$ decreases with the distance between $\params_i$ and $\params_j$, expressing the assumption that similar parameters yield similar objective values.

\fakepar{Posterior predictive distribution}
Given a dataset
\begin{equation}
\bodata_\boiter=\{(\params_1, \obj_1), \ldots, (\params_\boiter, \obj_\boiter)\},
\end{equation}
we denote the GP posterior conditioned on $\bodata_\boiter$ by $\mathcal{GP}_\boiter$.
Conditioning yields a Gaussian posterior predictive distribution at any query location $\params$ for both the latent objective $\bar{\obj}\!\left(\params \mid \bodata_\boiter \right)$ and a future noisy observation $\obj\!\left(\params \mid \bodata_\boiter \right)$,
\begin{equation} \label{eq:summarypredGPR}
\begin{aligned}
 	\bar{\obj}\!\left(\params \mid \bodata_\boiter \right) &\sim \mathcal{N}\!\left(\mu_\boiter(\params), \sigma^2_\boiter(\params)\right), \\
    \obj\!\left(\params \mid \bodata_\boiter \right) &\sim \mathcal{N}\!\left(\mu_\boiter(\params), \sigma^2_\boiter(\params) + {\sigma_\mathrm{n}}^2\right)\mathrm{.}
\end{aligned}
\end{equation}
Computing the predictive mean $\mu_\boiter(\params)$ and variance $\sigma^2_\boiter(\params)$ requires solving a linear system involving the $\boiter \times \boiter$ kernel matrix; direct methods scale cubically in $\boiter$.

\fakepar{Hyperparameters}
The prior GP is specified by a choice of mean and kernel family and the noise variance. In practice, these components depend on hyperparameters $\gphyp$ (\eg kernel lengthscales, signal variance, and ${\sigma_\mathrm{n}}^2$), which are typically estimated from data.
Given inputs $\Theta_\boiter$ and observations $\boldsymbol{J}_\boiter$, a standard approach is to maximize the (log) marginal likelihood $\mathrm{P}\!\left(\boldsymbol{J}_\boiter \mid \Theta_\boiter, \gphyp\right)$, i.e., maximum likelihood estimation, where $\mathrm{P}$ denotes probability.
Because BO often operates with few observations, especially early on,  maximum likelihood estimates can be poor.
A common remedy is to place a hyperprior $\mathrm{P}(\gphyp)$ and maximize the posterior, yielding a maximum a-posteriori estimate of $\gphyp$ proportional to
$\mathrm{P}\!\left(\boldsymbol{J}_\boiter \mid \Theta_\boiter, \gphyp\right)\,\mathrm{P}(\gphyp)$. \cite{rasmussen2006gaussian}

\subsection{Bayesian Optimization}  \label{sec:p1BOIntro}

BO is a sample-efficient stochastic black-box optimization method that approximates a function's global optimum without knowing the function's analytical form or gradients. It is effective for non-convex functions and noisy, expensive objective function evaluations. It is a well-known optimization algorithm that has been applied to a wide variety of problems in engineering, such as structural design, drug design/discovery, pharmaceutical product development, and new material design  \cite{garnett2022bayesian, candelieri2021gentle}. The reader is referred to \cite{shahriari2016taking} and \cite{garnett2022bayesian} for a more extensive introduction to BO.
\begin{algorithm}[ht]
\caption{Single-Objective Unconstrained BO}\label{algo:SOBO}
\begin{algorithmic}[1]
    \State \textbf{Input}: initial parameters $\params_1, \dots, \params_{\boiter_\mathrm{init}}$ 
    \State $\obj_1, ..., \obj_{\boiter_\mathrm{init}} \leftarrow$ run $\boiter_\mathrm{init}$ expensive-to-evaluate episodes with $\params_1, \dots, \params_{\boiter_\mathrm{init}}$  \label{algostate:initialSampling}
    \State $\bodata_{\boiter_\mathrm{init}} \leftarrow \{(\params_1,\obj_1 ),..., (\params_{\boiter_\mathrm{init}},\obj_{\boiter_\mathrm{init}} ) \}$
    \State $k \leftarrow k_\mathrm{init}$
    \Repeat \label{algostate:start-loop} 
       \State $\mathcal{GP}_\boiter \leftarrow $ Fit GP model of $J(\params)$ using $\mathcal{D}_\boiter$ with
       hyperparameter optimization \label{algostate:GPmodelling}  
    \State $\params_{\boiter+1} = \textrm{argmax}_{\paramSet}
    \alpha (\mathcal{GP}_\boiter)$ \label{algostate:accFunOpt}  
    \State $ (\obj_{\boiter+1})  \leftarrow$ run episode with $\params_{\boiter+1}$ \label{algostate:evaluation}  
    \State $\bodata_{\boiter+1} \leftarrow \bodata_{\boiter}  \cup \{(\params_{\boiter+1}, \obj_{\boiter+1})\}$ \label{algostate:dataupdate}  
    \State $\boiter \leftarrow \boiter+1$ 
    \Until{ stopping criterion is met, \eg $\boiter+1 \ge \boiter_{\mathrm{max}}$ } \label{algostate:end-loop}
    \State \Return approximated optimum $\hat{\params}^*$ 
    \label{algostate:return}
\end{algorithmic}
\end{algorithm}

Algorithm \ref{algo:SOBO} presents BO for a single-objective unconstrained optimization problem. First, during initial sampling, an initial data set $\bodata_{\boiter_\mathrm{init}}$ is created (Step \ref{algostate:initialSampling})
by evaluating multiple initial controller parameterizations $\params_1, \dots, \params_{\boiter_\mathrm{init}}$, where $\boiter_\mathrm{init} \geq 1$. %
The controller parameters are evaluated by running the expensive-to-evaluate closed-loop episode once for each parameterization to obtain noisy samples $\obj_1, \dots, \obj_{\boiter_\mathrm{init}}$.  

After the initial sampling, the main optimization loop is entered (Steps \ref{algostate:start-loop}-\ref{algostate:end-loop}). In each iteration $\boiter$, a probabilistic surrogate model, in this case, a GP regression model (see Sec. \ref{sec:p1GPRIntro}), is fitted to all past evaluations $\bodata_\boiter$ to approximate the unknown objective function $\obj(\params)$ (Step \ref{algostate:GPmodelling}). %

In Step \ref{algostate:accFunOpt}, the GP model is used to determine the parameterization to be evaluated next $\params_{\boiter+1}$. For this purpose, a so-called acquisition function $\alpha(\params)$ is defined. It estimates the utility of evaluating the objective function at a certain point in the parameter space. 
The parameterization, expected to contribute most to the progress of the optimization, is determined by maximizing the acquisition function. The acquisition function's main task is to balance exploration (sampling where little is known about the objective function) and exploitation (sampling where good performance is expected) using probabilistic GP predictions. One of the simplest acquisition functions is called upper confidence bound (UCB) (for minimization lower confidence bound (LCB)) where the parameter $\beta$ balances between exploration and exploitation:
\begin{equation}
\begin{aligned} \label{eq:acquisitionfun}
\params_{\boiter+1} &= \mathrm{argmax}_{\params \in \Theta} \quad \alpha_{\mathrm{LCB}} (\params), \\%\mathrm{with} 
\quad \alpha_{\mathrm{LCB}} (\params) &= -\left(\mu_k \left( \params \right) - \beta \sigma_k \left(\params\right) \right)
\end{aligned}
\end{equation}

The LCB acquisition function is large in areas where the cost function is expected to be small, i.e., small $\mu_k \left( \params \right)$, and/or the uncertainty is large, i.e., large $\sigma_k \left( \params \right)$. In Sec. \ref{sec:p2acquisitionfun}, we recommend using a more complex acquisition function without impactful hyperparameters such as $\beta$ in GP-UCB. Parameters $\boldsymbol{\theta}_{k+1}$ are then evaluated %
to obtain the corresponding noisy objective function value $J_{k+1}$ (Step~\ref{algostate:evaluation}). Afterward, the data set is augmented with the newly obtained information (Step~\ref{algostate:dataupdate}) and the next iteration starts.
After the maximum budget is reached or another stopping criterion is met, we estimate the best parameter value from the final GP model (Steps~\ref{algostate:end-loop} and~\ref{algostate:return}).

\begin{tcolorbox}[bluebox, title={Illustrative cart-pole example (Fig. \ref{fig:bo_intuition}) \hfill {\it open in} \href{https://github.com/Data-Science-in-Mechanical-Engineering/tunecontrol/blob/main/examples/standard_bo_for_controller_tuning.ipynb}{\faGithub}}]
\begin{center}
    \textit{\dots continuation of the example from Sec.~\ref{sec:p1problemstatement} \dots}
\end{center}

\smallskip
\fakepar{\circstep{4.} Optimize Control Parameters}
   Figure \ref{fig:bo_intuition} shows how BO is applied to the one-dimensional 
   cart-pole tuning problem defined in Sec.~\ref{sec:p1problemstatement}. 
   
   We follow the BO design choices outlined in Sec.~\ref{sec:p2vanillaBO}: zero mean, Gaussian likelihood, Matern 5/2 kernel, maximum a-posteriori optimization for hyperparameter optimization, and  max-value entropy search (MES).
Please refer to Sec.~\ref{sec:p2vanillaBO} on alternative BO design choices. 
   
    After three \emph{initial} random samples, the GP model predicts a lower cost function value near the lower bound of the domain. The uncertainty is also high in this region, as it has not yet been sampled. Consequently, the MES acquisition function identifies this area as the most beneficial for exploration. In \emph{BO Step 2}, a new sample at $\theta = -10$ is added to the dataset, updating the GP model. The following step focuses on exploitation, sampling near the current best observation. After another exploitation step, \emph{BO Step 5} demonstrates how, guided by MES, BO can escape a local predicted optimum and explore the region around $\theta = 4$. By \emph{BO Step 10}, the GP model predicts the optimum accurately, and the objective function is well estimated near the optimum, despite substantial noise in the samples.

    \medskip
    \fakepar{\circstep{5.} Deployment} The predicted optimal parameters $\hat{\params}^*$ can now be applied to the system.

\end{tcolorbox}

\section{Distinction from Other Controller Learning Paradigms: \textit{When to use BO?} } \label{sec:p1otherparadigms} %

Learning in control and robotics is a growing field that has produced many different methods for a wide variety of learning problems. In this section, we differentiate BO from other learning paradigms and highlight in which use cases BO is most useful. Specifically, we discuss alternative black-box optimization algorithms (Sec.~\ref{sec:p1otheroptimizers}), reinforcement learning (Sec.~\ref{sec:p1DRL}), and other learning-based control approaches (Sec.~\ref{sec:p1learningbasedcontrol}). %
For a detailed survey on data-efficient policy search in robotics we refer the reader to \cite{chatzilygeroudis2020survey}.

\begin{tcolorbox}[greybox, title={Key Messages}]

BO should be used instead of other black-box optimizers when the objective is expensive to evaluate and noisy (Sec.~\ref{sec:p1otheroptimizers}). 
BO should be preferred over deep reinforcement learning when limited system interactions are available and policies from classical control are sufficiently expressive. Furthermore, BO is applicable to non-Markovian systems and policies, can handle sparse rewards, and has less impactful hyperparameters (Sec.~\ref{sec:p1DRL}). 
BO can be used in combination with classical and learning-based control methods by optimizing their (hyper)parameters (Sec.~\ref{sec:p1learningbasedcontrol}).

\end{tcolorbox}

\subsection{Alternative Black-Box Optimization Algorithms} \label{sec:p1otheroptimizers}

The problem formulated in Sec. \ref{sec:p1problemstatement} can be addressed using other black-box optimization algorithms.
Examples include the well-known particle swarm optimization (PSO) or 
covariance matrix-adaptive evolutionary search (CMA-ES). 
\cite{panda2006comparison, artale2017comparison, chao2006comparison,mahesh2016performance, hendra2016comparison}.

It was shown on synthetic benchmarks (\eg \cite{leriche2021revisiting}), in %
machine learning (\eg \cite{turner2021bayesian}), and control and robotics, \eg \cite{calandra2016bayesian, stenger2022benchmark}  that BO is more sample-efficient than other black-box optimizers. However, alternative surrogate-based optimization algorithms such as global optimization via inverse distance (GLIS) weighting
can reach competitive performance to BO \cite{bemporad2020global}. Similarly, set membership global optimization (SMGO) achieved similar sample-efficiency as BO according to \citet{sabug2021smgo, catenaro2025automatic}.  
Other key advantages of BO over competing black-box optimizers are its ability to facilitate noisy objective function evaluations and the many extensions that have been proposed for more complex problem formulations (Sec.~\ref{sec:p2advanced_variants}) rendering it a practical framework for real-world problems. 

BO comes with a substantial amount of computational overhead because, at each iteration, a GP model is fitted to the data, and the acquisition function needs to be optimized. Therefore, BO is primarily applicable in cases where objective function evaluations are \emph{expensive} in terms of time or other resources. 

Some research has been conducted in combining BO with other black-box optimizers \cite{duivenvoorden2017constrained,pitra2016doubly}.
For example, the combination of BO and pattern search has shown great results in tuning for control \cite{stenger2022benchmark} and offers possibilities for future research.
 
\subsection{Reinforcement Learning} \label{sec:p1DRL}
 
Reinforcement learning (RL) is a learning paradigm in which an agent learns to make sequential decisions by interacting with an environment and receiving reward feedback.
In its most general form, RL seeks to optimize a policy that maximizes cumulative reward over time.
Two of the core challenges that make general RL difficult are \textit{(i)}~the typically high dimensionality of the policy parameter space and \textit{(ii)}~the sequential nature of the decision-making problem, where current actions influence future states and rewards.
 
Deep reinforcement learning (DRL) addresses expressive policy representation by parameterizing policies as neural networks with thousands or millions of parameters \citep{murphy2024reinforcement}.
This expressiveness enables exceptional asymptotic closed-loop performance on complex control tasks, often surpassing classical control approaches, \eg in vision-based robotic manipulation~\citep{kalashnikov2018scalable}.
However, the high-dimensional parameter space, combined with the inherent difficulty of sequential decision making, makes model-free DRL notoriously data-hungry, typically requiring millions of environment interactions for training.
Moreover, training success heavily depends on carefully engineered, application-dependent hyperparameters and reward shaping \citep{hwangbo2019learning}.
 
Optimizing controller parameters with BO (\cf \eqref{eq:single_objective}) can be understood as a special case of RL; specifically, episodic, model-free policy search~\citep{desisenroth2011survey}.
What fundamentally distinguishes BO-based controller tuning from general RL is the deliberate exploitation of prior control engineering knowledge.
This prior knowledge manifests in two ways that directly address the two core challenges above:

\begin{enumerate}[label=\arabic*), leftmargin=*, labelsep=0.5em]
\item \textit{A low-dimensional and structured parameter space.}
BO leverages well-established controller structures such as PID, LQR, or MPC that are known from control theory to be effective for specific classes of tasks.
These structures encode substantial domain knowledge into their architecture, so that only a small number of free parameters remain to be tuned.
This drastically reduces the search space from thousands or millions of neural network weights to typically fewer than ten interpretable parameters.
\item \textit{Transforming a dynamic problem into a static optimization problem.}
The use of established controller structures abstracts away the sequential decision-making aspect of the problem entirely.
Because the controller structure already handles the system dynamics the remaining task is to find the best \emph{static} parameter setting.
The dynamics are, in a sense, absorbed by the controller, and the optimization reduces to evaluating an episode-level cost function $\obj(\params)$ as a function of the controller parameters.
This is in stark contrast to general RL, where the policy must simultaneously learn to cope with the system dynamics and optimize long-term performance.
\end{enumerate}
 
Combining structured policies with the data-efficiency of BO enables effective learning with limited interaction time, placing BO-based controller tuning within the micro-data regime of RL~\citep{chatzilygeroudi2020survey}, where only tens of experiments may suffice instead of millions.
However, this efficiency comes at the cost of reduced asymptotic flexibility, as performance is inherently bounded by the expressiveness of the chosen controller structure.
Moreover, designing suitable controller structures often requires substantial domain expertise and engineering effort.

The black-box nature of BO naturally accommodates non-Markovian objectives, policies, and system dynamics, whereas standard RL formulations typically rely on Markovian assumptions.
BO can also handle sparse rewards and objectives defined over entire episodes without requiring modifications to the underlying algorithm.
Finally, the two paradigms are not mutually exclusive: BO can be used to optimize hyperparameters of RL algorithms~\cite{che2026efficient}, an approach commonly called AutoRL~\citep{parker2022automated}.

\subsection{Learning-Based Control} \label{sec:p1learningbasedcontrol}
We adopt the broad definition of learning-based control (LBC) as \emph{any method that uses machine learning in the design or operation of a control system}.
Under this definition, BO for controller tuning is itself a form of learning-based control: BO has been developed largely in the machine learning community and builds on probabilistic machine learning techniques, most notably GPR.
 
A variety of other LBC methods have gained significant popularity in recent years.
Reinforcement learning (discussed in Sec.~\ref{sec:p1DRL}) learns control policies from interaction data.
Learning-based model predictive control, in its many flavors (\eg \citep{coulson2019data,hewing2020learning}), combines data-driven models or data-driven predictions with receding-horizon optimization.
Model-based policy search methods (\eg \citep{deisenroth2011pilco,deisenroth2013gaussian}) first learn a dynamical model from data, which then serves as the basis for obtaining controller parameters.
While model-based methods can be beneficial in terms of data efficiency, they strongly rely on the learned model accurately capturing the true dynamics, which is itself a challenging problem~\citep{doerr2017optimizing}.
 
The use of data and learning in control is, however, not new and has deep roots in classical control theory.
Adaptive control methods seek to adjust model or controller parameters online in response to data~\citep{annaswamy2021historical}.
Iterative learning control (ILC) improves performance over repeated task executions \citep{bristow2006survey}.
Data-driven control methods such as virtual reference feedback tuning (VRFT)~\citep{campi2002virtual} directly compute controller parameters from experimental data.
 
What distinguishes BO within the broader LBC landscape is its particular combination of characteristics: it treats the closed-loop system as a black box, leverages controller structure to reduce the problem to optimizing on the order of tens of parameters, and is specifically designed for settings where each evaluation is expensive, for example, a hardware experiment or a high-fidelity simulation.
 
Crucially, most LBC methods involve parameters or hyperparameters that must themselves be tuned.
For example, learning-based MPC requires the specification of weighting matrices, prediction horizons, terminal costs, and regularization parameters.
DRL depends on learning rates, network architectures, and reward shaping.
Even adaptive control schemes have adaptation gains and design parameters.
This makes BO an orthogonal and complementary tool that can be applied \emph{on top of} other LBC methods to tune their hyperparameters.
Indeed, BO is already widely used for hyperparameter tuning in machine learning more broadly~\citep{snoek2012practical,turner2021bayesian}.
In the control domain, successful examples of BO tuning learning-based controllers include learning-based MPC~\citep{frohlich2021model}, approximate MPC~\citep{hose2024fine}, and extremum seeking controllers~\citep{charkabarty2022extermum}.
BO thus provides an orthogonal meta-optimization layer for tuning LBC methods.

\part{Review}

Part B addresses advanced users of BO and researchers developing BO-algorithms for control and robotics. First, we review design choices in unconstrained single-objective BO (``vanilla BO'') and give recommendations on starting points for BO hyperparameters (Sec.~\ref{sec:p2vanillaBO}). Building on this, we review advanced challenges that occur in controller and robotics tuning, formally define the respective optimization problems and explain how BO can solve them (Sec.~\ref{sec:p2advanced_variants}). 
Lastly, we systematically survey a total of 110 papers that apply BO to robotics and control on hardware (Sec.~\ref{sec:p2applications}).

\section{Design Choices in Vanilla Bayesian Optimization} \label{sec:p2vanillaBO}

Building on Sec.~\ref{sec:p1BOandGPRIntro}, this section reviews the main design choices for single-objective BO and offers recommendations for practitioners. These recommendations draw on both prior work and the authors’ experiences. However, a systematic evaluation on representative control- and robotics-specific benchmarks is currently still lacking in the literature (see Sec.~\ref{sec:p3futureissues}), so these recommendations should be interpreted as an informed starting point rather than definitive conclusions.

\begin{figure*}[ht]
    \centering
    \includegraphics{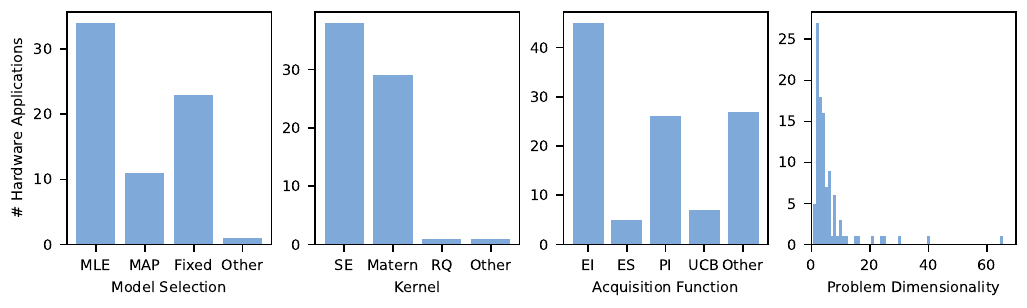}
    \caption{Statistics on the model selection method, kernel (see Sec. \ref{sec:p2GPRmodel}), acquisition function (see Sec. \ref{sec:p2acquisitionfun}) and problem dimensionality of control and robotics paper that use BO for online parameter tuning on hardware. Papers are selected according to the review method described in Sec. \ref{sec:p2reviewmethod}. Total number of papers: 110. Papers where the respective information is missing are excluded in this illustration.}
    \label{fig:statistics}
\end{figure*}

The two key components of BO are a probabilistic model of the objective function and an acquisition function that, based on this model, will decide the next query location.
In the following, we will first review the key aspects of the model (Sec.~\ref{sec:p2GPRmodel}), mainly focusing on GPs, and then discuss different acquisition functions (Sec.~\ref{sec:p2acquisitionfun}). 
For quantitative analysis of the most popular design choices, we examined 110 papers that applied BO directly on hardware according to the criteria discussed in Sec.~\ref{sec:p2reviewmethod}. Results are summarized in Fig.~\ref{fig:statistics}.  

\begin{tcolorbox}[greybox, title={Key Messages}]

As a starting point, we recommend the following design choices in BO: A GPR model with homoscedastic Gaussian likelihood, a Matérn or SE-kernel with ARD, zero or constant mean with standardized observations, maximum a-posteriori hyperparameter optimization (Sec.~\ref{sec:p2GPRmodel}), and max-value entropy search for particularly noisy objectives and logEI for high-dimensional and low-noise settings (Sec.~\ref{sec:p2acquisitionfun}). Furthermore, we recommend $d$ plus a few, \eg $d+4$, initial data points, scale input and output data, and select the final solution by minimizing the expected cost over all evaluated parameterizations (Sec.~\ref{sec:p2otherpractical}). 

\end{tcolorbox}

\subsection{Probabilistic Surrogate Model} \label{sec:p2GPRmodel}

\fakepar{Model Class}
BO has its name from relying on a probabilistic surrogate model. At each iteration, the model is updated with an additional observation and used to predict the unknown objective function, see \eqref{eq:summarypredGPR}. Examples of popular BO surrogate models are GPs (\eg \cite{balandat2020botorch}) (also known as Kriging), random forest (\eg \cite{lindauer2022smac3}), and tree-structured parzen estimators (TPE) (\eg \cite{watanabe2023tree}). With few exceptions, %
the vast majority of papers for controller tuning use GPs as surrogate models. In fact from the 110 reviewed hardware applications (see Sec.~\ref{sec:p2reviewmethod}) only \citet{bellegarda2024quadruped} used TPEs instead of GPs. Furthermore, most of the extensions in Sec. \ref{sec:p2advanced_variants} are based on GPs. 
This popularity of GPs as surrogate models may be explained by the analytical posterior computations~\cite{rasmussen2006gaussian} and appealing theoretical properties of GPs in the context of BO~\cite{srinivas2009gaussian}. 
Additionally, popular BO implementations like, BoTorch \cite{balandat2020botorch},
mainly rely on GPs as surrogate models.

GPs work best for real-valued input spaces and small data sets as the computational complexity of inference scales cubically with the data set size \cite{rasmussen2006gaussian}. Many works have been proposed to address non-real-valued input spaces (Sec.~\ref{sec:p2decisionspaces}) and to enable scalable GP regression \cite{liu2020gaussian}. Still, random forests and TPEs can handle categorical and integer variables out of the box \cite{bliek2023benchmarking}. Furthermore, they scale better with larger problem dimensions \cite{bliek2023benchmarking}. Empirically, they perform competitive to GP-based BO depending on the problem setting \cite{bliek2023benchmarking} although exhaustive benchmarks on control and robotics problems are lacking.     
Furthermore, in recent years, using Bayesian neural networks for BO has shown promising results on tasks with complex input correlations \cite{li2024study,kristiadi2023promises,brunzema2024variational}.
The ability to learn input correlations could also prove beneficial in controller tuning; however, apart from a few exceptions \cite{chakrabarty2022optimizing}, this direction has remained largely unexplored. In what follows, we concentrate on GPs as surrogates, though practitioners should keep in mind that competitive alternatives  exist.

\begin{tcolorbox}[greenbox, arc=2pt, left=3pt, right=3pt, top=3pt, bottom=3pt]
\textbf{Recommended starting point:}  GPR. It is the most established method in control and robotics and has been extended to many advanced settings (see Sec.~\ref{sec:p2advanced_variants}). 
\end{tcolorbox}

\fakepar{Likelihood / Observation Model}
The likelihood model encodes assumptions on the noise distribution of the black-box function (see \eqref{eq:likelihood}).
The most popular observation model for controller tuning is homoscedastic Gaussian distributed independent noise.
Homoscedastic means that the noise distribution has the same variance independent of the parameters. Homoscedastic Gaussian noise has two major advantages: It is analytically tractable in the GP framework and has only one hyperparameter, the variance of the observation noise.

However, the homoscedastic assumption may not hold in practice.
Thus, heteroscedastic, \ie parameter-dependent, noise has been considered in controller tuning problems \cite{ariizumi2017mutiobjective,guzman2020heteroscedastic,guzman2022bayesian} and robot learning, \eg \cite{kuindersma2012variational}.  
Heteroscedastic noise models typically have more parameters and therefore, more data points tend to be needed to accurately estimate the noise level. 

One disadvantage of the Gaussian likelihood is that outliers far away from the center of the distribution are very unlikely to be observed. Therefore, outliers heavily deteriorate prediction accuracies in GPs with Gaussian likelihood. To address this, \eg \citet{martinez2018practical}, propose an outlier detection method based on the Student's-t-likelihood. Student's-t-distributions are heavier-tailed than Gaussian distributions, \ie they are less susceptible to outliers.  
Optimization with a Student's-t-likelihood was also applied for Kalman filter tuning in \cite{bertipaglia2022two, chen2019kalman} and controller tuning \cite{stenger2020robust,sasaki2020bayesian}. However, a Student's-t-likelihood has an additional hyperparameter compared to a Gaussian likelihood and thus tends to need more data to be fitted accurately.

Another special case arises when the objective function evaluation is deterministic. However, when changing parameters slightly, we get a noise-like perturbation of the trend we are interested in. 
These perturbations may be lumped in the noise model of the GP.
This behavior is sometimes called \emph{deterministic noise} \cite{forrester2006design}. 
Deterministic noise needs to be addressed explicitly in the acquisition function 
because querying the objective function twice at the same location does not result in new information.

\begin{tcolorbox}[greenbox, arc=2pt, left=3pt, right=3pt, top=3pt, bottom=3pt]
\textbf{Recommended starting point:}  Homoscedastic Gaussian noise. It is the most popular model with the fewest hyperparameters. However, it is essential to distinctly treat outliers and deterministic noise should they occur.
\end{tcolorbox}

\fakepar{Kernel}
As discussed in Sec.~\ref{sec:p1GPRIntro} the kernel function expresses the correlations of the output given the input and can express, for example, smoothness or periodicity.
The most commonly used kernels are the squared exponential (SE) and kernels from the Mat\'ern family (cf. Fig. \ref{fig:statistics}).
For both, the correlations between two experiments simply decrease with increasing distance in the parameter space. They differ in their smoothness assumptions. The SE kernel assumes functions are in $\mathcal{C}^\infty$ where the Mat\'ern kernel can be parameterized to be less smooth. 
Comparative studies on controller tuning \cite{snoek2012practical,leriche2021revisiting, stenger2022benchmark} show ambiguous results.

Kernels can be further categorized in isotropic and anisotropic, \ie same length scale for all dimensions, or different length scales for all dimensions (also called automated relevance detection (ARD)). In general, an ARD kernel is more expressive and can better model the objective function landscape. However, it comes with more hyperparameters, and therefore, more data points are needed to fit the model. Using a hyperprior (see paragraph on model selection and hyperparameters) can effectively address this issue. 

Additionally, kernels can be characterized as stationary and non-stationary.
Stationary kernels model only the relative distance between the inputs. 
In contrast, non-stationary kernels also explicitly depend on input locations and therefore allow for higher flexibility.
Usually, stationary kernels are used for BO. However, non-stationary kernels have shown promising results \cite{martinezcantin2017bayesian,martinezcantin2019funneled}.

The kernel choice is a powerful way to include advanced prior domain knowledge. 
For example \cite{marco2017design} developed a specific kernel for the LQR controller. The studies \cite{antonova2017deep, rai2018bayesian,antonova2020bayesian} design domain-specific kernels by leveraging simulation results. This way the search space can be greatly reduced, \eg from 50 to 2 parameters \cite{rai2018bayesian}. \cite{jaquier2020bayesian, jaquir2022geometry} also use a domain-specific kernel for robot learning involving Riemannian manifolds.

A different approach to encode prior knowledge on the objective function are shape constraints. This way assumptions about \eg monotonicity or convexity, of the objective function can be enforced \cite{agrell2019gaussian, brunzema2022controller}.

\begin{tcolorbox}[greenbox, arc=2pt, left=3pt, right=3pt, top=3pt, bottom=3pt]
\textbf{Recommended starting point:}  Mat\'{e}rn  or SE kernel with ARD. Both are fairly general and well established.   
\end{tcolorbox}

\fakepar{Mean Function}
As discussed in Sec.~\ref{sec:p1GPRIntro}, the GP prior mean function heavily influences the extrapolation behavior of the GP. As a result, it is an obvious way to include prior knowledge in the optimization, \eg \cite{lu2021bayesian}.
Usually in controller tuning or robot learning a constant or zero mean is chosen because no additional prior knowledge is available.
In fact, only three \cite{pautrat2018bayesian,culha2020learning,tang2022borderline} of the 110 reviewed hardware application papers (see Sec.~\ref{sec:p2reviewmethod}) report a mean different from constant or zero. Note that a zero mean should only be used if the observations are scaled to have zero mean (see Sec.~\ref{sec:p2otherpractical}).

While constant or zero mean functions are simple and widely used, they impose a very limited prior structure on the objective. In particular, far away from observed data, the GP posterior mean tends to revert to this constant baseline and therefore cannot represent global trends in the objective function. This limitation has motivated the use of more expressive mean functions in BO. For example, the benchmark by \cite{leriche2021revisiting} suggests that a quadratic prior mean function can be promising on synthetic functions. However, this finding could not be confirmed for deterministic controller tuning tasks \cite{stenger2022benchmark}. Still, \cite{tang2022borderline} used second-order polynomials as mean functions in a real-world engine calibration task. 
Blind kriging \cite{joseph2008blind} is one way to automatically find the prior mean functions in GPs. %
Some preliminary work on blind kriging in BO \cite{mai2022improved,stenger2019machine} has shown some promising results that may be worth exploring in a control and robotics setting.

If low-fidelity (\eg simulator) data is available, it can be leveraged to construct the GP's prior mean function, \eg \cite{lu2021bayesian}. However, the correct prior model, \ie the correct simulation parameters may not always be known. This is why  \citet{pautrat2018bayesian} proposed a method to choose the best prior mean function from a set of prior models constructed with differently parametrized simulations automatically. Experimental results from related systems can also be beneficial. In fact, \cite{culha2020learning} have shown that reusing the posterior mean of a related system as the prior mean of a slightly different robot can be beneficial.

\begin{tcolorbox}[greenbox, arc=2pt, left=3pt, right=3pt, top=3pt, bottom=3pt]
\textbf{Recommended starting point:} Zero or constant mean in combination with standardized observations. In most cases no additional prior knowledge is available. 
\end{tcolorbox}
 
\fakepar{Model Selection \& Hyperpriors}
Fig.~\ref{fig:statistics} summarizes how GP hyperparameters are commonly selected in practice. Three main approaches are used: fixing them a priori, estimating them online by maximum-likelihood estimation (MLE), or estimating them using maximum-a-posteriori (MAP) estimation, see Sec.~\ref{sec:p1GPRIntro}.

In practice, fixing GP hyperparameters a priori can be difficult and a poor choice can lead to inefficient or unsuccessful optimization. 
Fortunately, many applications demonstrate that BO can still perform well without prior hyperparameter knowledge by adapting the GP hyperparameters online by maximizing the marginal likelihood as new observations become available.

However, the MLE approach can result in hyperparameters that contradict engineering domain knowledge, for example, extremely small or large observation noise. An effective tool to introduce mild domain knowledge without fixing hyperparameters is to use MAP. This way, for example, length scales and observation noise can be roughly bounded to realistic ranges without precisely fixing them. 
Commonly used hyperpriors for length scales are box-hyperpriors \eg \cite{stenger2023automatic}, in a unit cube domain, Gamma-hyperpriors (\eg \cite{gabler2022bayesian}), and log-normal hyperprior (\eg \cite{hvarfner2024vanilla}). To save computational overhead for hyperparameter optimization, one can consider only retraining after, \eg 10 iterations \cite{polonio2022bayesian}. 

From a theoretical point of view, decreasing the length scale over the course of optimization seems promising \cite{berkenkamp2019noregret}. However, for high dimensional problems this is not practical. In fact, pushing hyperparameter optimization toward larger length scales has empirically demonstrated great potential \cite{hvarfner2024vanilla,xu2025standard} even for higher-dimensional tasks.
A suitable range for the noise hyperprior can be determined by repeatedly evaluating an initial parametrization before optimization.

\begin{tcolorbox}[greenbox, arc=2pt, left=3pt, right=3pt, top=3pt, bottom=3pt]
\textbf{Recommended starting point:} Maximum a-posteriori estimation effectively biases hyperparameters to sensible ranges while requiring little prior knowledge.
\end{tcolorbox}

\subsection{Acquisition Function} \label{sec:p2acquisitionfun}

The maximum of the acquisition function determines the location of the next sample. It resolves the trade-off between exploration and exploitation \eqref{eq:acquisitionfun}. We refer to Sec. \ref{sec:p2advanced_variants} for acquisition functions addressing advanced challenges in BO. 

Acquisition functions can be divided into four categories: 
(i) \emph{Optimistic}, \eg upper confidence bound (UCB) \cite{auer2002confidence},
(ii) \emph{Improvement-based}, \eg probability of improvement (PI), expected improvement (EI) \cite{jones1998efficient}), log expected improvement (logEI) \cite{ament2023unexpected}
(iii) \emph{Information theoretic}, \eg entropy search \cite{hennig2012entropy}, predictive entropy search \cite{hernandezlobato2014predictive}, max-value entropy search (MES) \cite{wang2017max}, joint entropy search \cite{hvarfner2022joint,tu2022joint}, and local entropy search \cite{stenger2026local}
(iv) \emph{Thompson sampling} (\eg \cite{agrawal2013thompson})

We refer to \cite{garnett2022bayesian} for a detailed explanation and broader introduction to acquisition functions. In principle, all of these acquisition functions can be used for controller tuning tasks. This is reflected by the diverse usage statistics Fig.~\ref{fig:statistics}. 

From an empirical point of view, it is inconclusive which acquisition functions perform best on a wide range of controller tuning tasks. Different acquisition functions have
been compared for one specific application, \eg in \cite{calandra2016bayesian,neumannbrosig2019data,vonrohr2018gait}, from which we cannot draw general conclusions. In a controller tuning benchmark on ten deterministic simulative controllers \cite{stenger2022benchmark}, it was shown that EI and MES performed similarly on average, with UCB performing slightly worse. The ambiguity of the results, \ie the dependency of the preferred BO setup on the test case, supports the findings of \cite{turner2021bayesian}, where BO ensembles using various acquisition functions and surrogate models performed best.

From a practical point of view, it needs to be considered that UCB has an influential hyperparameter $\beta$ that balances exploration and exploitation.
While there exist theory on how $\beta$ should be chosen to guarantee convergence \cite{srinivas2009gaussian,chowdhury2017kernelized}, it can in practice result in too much exploration and undesirable empirical performance.
To still have a $\beta$ that follows insights from theoretical investigations, \ie it should logarithmically increase over time. \citet{kandasamy2015high} propose to set $\beta = 0.2 d \log (2k)$.
However, such a heuristic is rarely used in practice and it is currently more popular to set $\beta$ to a fixed value. %

A recent extension to EI, logEI \cite{ament2023unexpected}, has shown great performance even in high-dimensional problems \cite{hvarfner2024vanilla}. However, care must be taken when using the expected improvement in the noisy setting. The standard formulation originally developed for the deterministic case should not be used in the noisy case see, \eg \cite{letham2019constrained}. However, the popularity of standard EI in control and robotics (Fig.~\ref{fig:statistics}) suggests that in practice it may still be fine to use standard EI even in noisy cases.

Information-theoretic acquisition functions do not have influential hyperparameters except for approximation accuracies and can be used in the noisy and noiseless case. However, with the exception of MES, they are usually rather expensive to evaluate and %
special care needs to be taken in high-dimensional optimization problems.

To circumvent a fixed acquisition function choice, different approaches were developed to automatically select GP-acquisition functions \eg  \cite{hoffman2011portfolio,vasconcelos2022self,benjamins2022pi}. However, those methods are almost never applied in control and robotics. One exception is the application of GP-HEDGE \cite{hoffman2011portfolio} to robot learning \cite{daniel2014active} where it did not achieve a performance increase.

Above, we discussed myopic, i.e., one-step optimal, acquisition functions because they are mainly used in control and robotics. Non-myopic acquisition functions (see, \eg \cite{yue2020why}) show great potential at the expense of more computational effort.
Additionally, the acquisition function can be used to include prior user beliefs in the optimization, \eg \cite{hvarfner2022pi}.

Any acquisition function has generally more than one local maximum. Therefore, in practice, globally searching algorithms, \eg multi-start gradient
descent or evolutionary algorithms should be used to optimize the acquisition function.

\begin{tcolorbox}[greenbox, arc=2pt, left=3pt, right=3pt, top=3pt, bottom=3pt]
\textbf{Recommended starting point:} Max-value entropy search for noisy and low-dimensional problems. It has no hyperparameters apart from approximation accuracies and works in noisy and noiseless cases with an acceptable amount of overhead.   
logEI \cite{ament2023unexpected} for high-dimensional problems with low noise. It has shown exceptional performance in high dimensions in combination with high-lengthscale hyperpriors - see also Sec.~\ref{sec:p2high_dim}
\end{tcolorbox}

\subsection{Other Practical Consideration} \label{sec:p2otherpractical}

BO can benefit from \textbf{initial data points}. The initial data set can be uniformly random or, to reduce randomness and cover the whole search domain, generated using space-filling sampling (\eg Sobol-sequences or latin hypercube sampling). Exceptions are cases where cautious or safe exploration is required.
\citet{leriche2021revisiting} have shown that a small initial budget of $d+4$ samples performs better than larger initial budgets for most of the examined synthetic objective functions. Results in \cite{stenger2022benchmark} suggest that for deterministic tuning tasks $d+1$ random initial samples may be sufficient for BO to outperform baseline optimization algorithms.  

To avoid numerical issues in the GP model and acquisition function optimization and decrease the problem-dependence of hyperparameter selection it is common practice to use \textbf{data scaling} \cite{golovin2017google}. Specifically, we recommend scaling the input domain to the unit cube and normalize all outputs to zero mean and standard deviation of one.

A \textbf{stopping rule} (Step 11 in Alg. \ref{algo:SOBO}) defines when the optimization is stopped. It is common practice to use a fixed number of evaluations or to stop the optimization when performance increase is sufficient. Alternative stopping rules can be derived from the confidence bounds of the GP or the value of the acquisition function. Promising results on estimated model-based regret have been shown by \citet{wilson2024stopping, stenger2026local}. 

The naive way to \textbf{select the final solution} $\hat{\params}^*$  (Step 12 in Alg. \ref{algo:SOBO}) would be to take the parametrization with the best observed objective function value. However, due to the noisy objective function evaluations, this best observation does not have to occur for the best parametrization. Instead, we recommend selecting the evaluated parametrization with the lowest posterior mean.

\begin{tcolorbox}[greenbox, arc=2pt, left=3pt, right=3pt, top=3pt, bottom=3pt]
\textbf{Recommended starting point:} Use $d+4$ initial data points, scale input and output data, and select the final solution by minimizing the expected cost over all evaluated parameterizations
\end{tcolorbox}

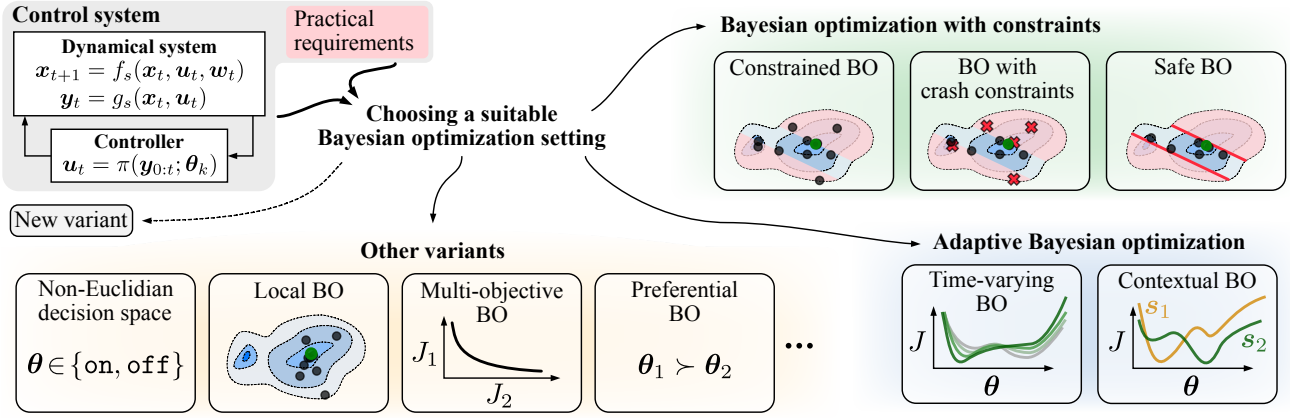
\begin{figure*}
    \centering
    \begin{tikzpicture}
        \node[anchor=south west,inner sep=0] (image) at (0,0) 
              {\includegraphics[]{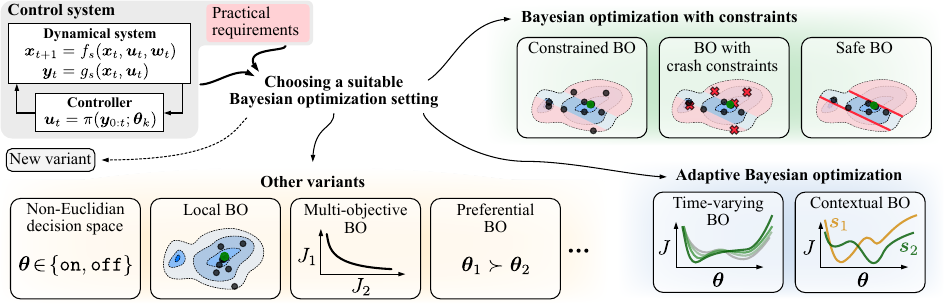}};
        
        \node[anchor=south west, inner sep=0, 
              minimum width=2.1cm, minimum height=1.7cm,
              fill=none] 
              at (0.25,0.14) {\hyperref[sec:p2decisionspaces]{\phantom{\rule{2.2cm}{1.7cm}}}};
        
        \node[anchor=south west, inner sep=0,
              minimum width=2.1cm, minimum height=1.7cm,
              fill=none] 
              at (2.8,0.14) {\hyperref[sec:p2high_dim]{\phantom{\rule{2.2cm}{1.7cm}}}};
        
        \node[anchor=south west, inner sep=0,
              minimum width=2.1cm, minimum height=1.7cm,
              fill=none] 
              at (5.4,0.14){\hyperref[sec:p2MultObj]{\phantom{\rule{2.2cm}{1.7cm}}}};
        
        \node[anchor=south west, inner sep=0,
              minimum width=2.1cm, minimum height=1.7cm,
              fill=none] 
              at (7.9,0.14){\hyperref[sec:p2preferential]{\phantom{\rule{2.2cm}{1.7cm}}}};
        
        \node[anchor=south west, inner sep=0,
              minimum width=2.1cm, minimum height=1.7cm,
              fill=none] 
              at (12.,0.25) {\hyperref[sec:p2timevarying]{\phantom{\rule{2.2cm}{1.7cm}}}};
        
        \node[anchor=south west, inner sep=0,
              minimum width=2.1cm, minimum height=1.7cm,
              fill=none] 
              at (14.6,0.25){\hyperref[sec:p2contextualbo]{\phantom{\rule{2.2cm}{1.7cm}}}};
        
        \node[anchor=south west, inner sep=0,
              minimum width=2.1cm, minimum height=1.7cm,
              fill=none] 
              at (9.5,3.1) {\hyperref[sec:p2Constrnt]{\phantom{\rule{2.2cm}{1.7cm}}}};
        
        \node[anchor=south west, inner sep=0,
              minimum width=2.1cm, minimum height=1.7cm,
              fill=none] 
              at (12.1,3.1) {\hyperref[sec:p2crash]{\phantom{\rule{2.2cm}{1.7cm}}}};
        
        \node[anchor=south west, inner sep=0,
              minimum width=2.1cm, minimum height=1.7cm,
              fill=none] 
              at (14.67,3.1) {\hyperref[sec:p2safeBO]{\phantom{\rule{2.2cm}{1.7cm}}}};
        
    \end{tikzpicture}
    \caption{Overview of BO extensions for controller tuning and robot learning tasks presented in Sec. \ref{sec:p2advanced_variants}. They allow various practical requirements of controller tuning tasks to be formalized and solved with BO. \hfill\textit{(Pictograms are clickable)}}
    \label{fig:overview}
\end{figure*}

\section{Advanced Variants} \label{sec:p2advanced_variants} %

In this section, we review common challenges that frequently occur in BO for control and robotics, and we present advanced BO formulations to address these. A summary is provided in Fig.~\ref{fig:overview}. 

\begin{tcolorbox}[greybox, title={Key Messages}]
The BO framework offers many extensions suitable for advanced problem settings in controller tuning (see Fig.~\ref{fig:overview}.) 
\end{tcolorbox}

\subsection{Constrained BO } \label{sec:p2Constrnt} 

\begin{tcolorbox}[methodbox]
\textbf{Goal:} Ensure final solution is feasible

\medskip
\textbf{Example:} Maximum overshoot is below some admissible maximum $\boldsymbol{g}_\mathrm{max}$

\medskip
\textbf{Available information:}
\begin{itemize}[left=10pt, itemsep=-3pt, topsep=1pt]
\item Noisy samples of objective $J_{k}$
\item Noisy samples of unknown constraint function~$\boldsymbol{g}_k$
\end{itemize}
\end{tcolorbox}

\fakepar{Problem Statement} Formulating additional constraints is likely necessary to ensure the practical feasibility of the tuned controller. For example, when minimizing the energy consumption of a reference tracking controller, the maximum tracking error should be constrained. Otherwise, the optimal solution would be to set all controller gains to zero. A constrained optimization problem can be written as follows: %
\begin{align}
    \params^* =  \argmin{\params \in \mathbb{R}^d}\qquad   &\mathbb{E}\left[ J(\theta) \right] \\
		\mathrm{s.t. }\qquad   & \params_{\mathrm{min}} \le \params \le \params_{\mathrm{max}} \\
        \quad &\mathrm{P}(\boldsymbol{g}(\params) \le \boldsymbol{g}_\mathrm{max}) \geq \alpha \label{eq:p2unknownconstraints} \\ 
        \quad &\boldsymbol{h}(\params) \label{eq:p2knownconstraints} \le \boldsymbol{h}_\mathrm{max} 
\end{align}
\textit{Unknown constraints} \eqref{eq:p2unknownconstraints} can only be evaluated by querying the black-box experiment. Thus, the constraint function $\boldsymbol{g}(\params)$ 
is calculated from the time-domain behavior in the same way as the objective function. Due to the potential stochasticity of the closed-loop system, a probabilistic constraint formulation has to be used, i.e., the threshold $\boldsymbol{g}_\mathrm{max}$ must not be exceeded with probability $\alpha$. Arbitrary constraint formulations can be chosen. Practical examples include not exceeding a critical process temperature. Note that the unknown constraints can also be binary. In contrast, \textit{known constraints} $\boldsymbol{h}(\params)$ \eqref{eq:p2knownconstraints} %
can be evaluated analytically without querying the closed-loop system. Examples of known constraints are analytical stability requirements \cite{dorschel2021safe} or box constraints. The search domain $\paramSet$ (\cf \ref{sec:p1problemstatement}) is defined in terms of the known constraints:
    $\paramSet = \left\{ \params \in \mathbb{R}^d | \params_{\mathrm{min}} \le \params \le \params_{\mathrm{max}} \land  
        \boldsymbol{h}(\params)  \le \boldsymbol{h}_\mathrm{max} \right\}$. 
In contrast to safe BO (cf. Sec. \ref{sec:p2safeBO}), constrained BO explicitly allows constraint violations during optimization. Only the final solution is required to be feasible. The main challenge of constrained BO is to suggest feasible parameters and yield an improvement of the objective function value.

\fakepar{Methods \& Applications}
\emph{Known constraints} can directly be included in the acquisition function optimization. \emph{Unknown constraints} are more challenging. Usually, an additional GPR model is formulated to estimate $\boldsymbol{g}(\params)$. In the case of binary constraints, this is a classification model. 
Using these additional models, probabilistic predictions of constraint violations can be obtained and  included in the acquisition function.

Numerous acquisition functions were adapted to the constrained case and applied to controller and robotics tuning in hardware experiments (see table \ref{tab:hardware_extensions}). The expected improvement criterion can be extended to the constrained case (\eg \cite{gardner2014bayesian, gelbart2014bayesian}) by multiplying the expected improvement with the probability of feasibility. This approach is used in \cite{khosravi2023safety} to optimize a CNC machine and in \cite{marco2021robot} for quadruped control. Constrained expected improvement (EIC) requires an initial feasible point for optimization. Additionally, EI and EIC are not suitable for the noisy case without modifications, \eg \cite{letham2019constrained}.
These shortcomings are, for example, alleviated by constrained max-value entropy search (CMES) \cite{perrone2019constrained,takeno2022sequential}, a constrained extension of the information-theoretic MES \cite{wang2017max}. A controller tuning hardware application of CMES can be found, for example, in \cite{stenger2023automatic}. 
Most constrained BO frameworks can also address binary constraints (cf., \eg \cite{perrone2019constrained, gelbart2014bayesian}). 

A special case of constrained BO allows a limited failure budget, \eg \cite{marco2020excursion,xu2022vabo} with application to robotics  \cite{yu2022learning}. In this setting, only a limited number of constraint violations are allowed during optimization. This can be seen as an intermediate case between safe and constrained BO (see \ref{sec:p2safeBO}). 

\subsection{Safe BO}%
\label{sec:p2safeBO}

\begin{tcolorbox}[methodbox]
\textbf{Goal:} Final solution and all samples have to be feasible

\medskip
\textbf{Example:} Maximum overshoot is always below $\boldsymbol{g}_\mathrm{max}$ 

\medskip
\textbf{Available information:}
\begin{itemize}[left=10pt, itemsep=-3pt, topsep=1pt]
\item Noisy samples of objective $J_{k}$
\item Noisy samples of unknown constraint function~$\boldsymbol{g}_k$
\end{itemize}
\end{tcolorbox}

\fakepar{Problem Statement} 
When applying BO to safety-critical applications, safety constraints arise. Violating these safety constraints can lead to damage to the experimental setup. One practical example is the reactor temperature never exceeding a critical threshold in chemical processes. Therefore, in safe BO, the objective function $J$ should be optimized without violating any safety constraint in any iteration of the optimization process: 
\begin{equation}
    \label{eq:safemaxproblem}
    \begin{aligned}
    & \params^* = \argmin{\params \in \paramSet}{\mathbb{E}\left[ J(\theta) \right]}, \\
    & \textrm{subject to} \ \boldsymbol{g}_\boiter \leq \boldsymbol{g}_{\boiter, \mathrm{max}} \ \forall \boiter = 1, ..., \maxboiter,
    \end{aligned}
\end{equation}
where $\boldsymbol{g}_\boiter$ is the noisy realization of $\boldsymbol{g}(\params)$. Achieving safety with known constraints \eqref{eq:p2knownconstraints} is straightforward, because we can ensure before the experiment that chosen parameters are safe.    
The main challenge in safe BO is to find a set of safe decisions in each iteration and trade-off expanding the safe set while finding the optimum in the safe set. 

\fakepar{Methods \& Applications}
The general structure of %
safe BO algorithm is to determine the safe set, optimize the acquisition function in the safe set, and finally query the next point. One key assumption is that in the first iteration, an initial safe parameter set is given. Different safe BO methods introduce different variants of the determination of the safe set and vary in their exploration strategy. 

\citet{sui2015safe} introduced the most popular approach for identifying the safe set, which relies on confidence bounds.
The assumptions in this setting are that the target function is from a Reproducing Kernel Hilbert Space (RKHS), is Lipschitz continuous, and observations are perturbed with subgaussian noise. With these assumptions, one can ensure with high probability the unknown function is inside the confidence bounds, which can be defined as $b(\params) = \mu(\params) \pm \beta \sigma(\params)$, where $\beta \ge 0$ is a parameter that determines the width of the confidence bounds. Different theoretical contributions derive $\beta$ \cite{abbasi2013online, srinivas2009gaussian, chowdhury2017kernelized,molodchyk2024towards}. However, these derivations are often conservative and rely on a known or estimated \cite{tokmak2025safe} RKHS norm bound and a correct choice of the kernel. A detailed discussion on using confidence bounds in SafeOpt can be found in \cite{fiedler2021practical, fiedler2024on}. %
In most practical applications, %
$\beta$ is set to a constant heuristic value, \eg $\beta = 2$ \cite{berkenkamp2016safe, berkenkamp2023bayesian, konig2021safe, wischnewski2019model} or $\beta = 3$ \cite{baumann2021gosafe, dorschel2021safe}. While such heuristics often invalidate theoretical guarantees, they may still yield useful ``cautious'' behavior in practice. Furthermore, safety also highly depends on the choice of the hyperparameters of the GP. Using a Matern kernel with $\nu = 3/2$ and short lengthscales \cite{berkenkamp2016safe, baumann2021gosafe, wischnewski2019model, weber2021safe} is most common. \citet{wischnewski2019model} report safety violations %
due to the misspecification of GP hyperparameters. Consequently, many safe BO works have reported fixed GP-hyperparameters, \eg \cite{schillinger2017safe,berkenkamp2016safe,widmer2023tuning,baumann2021gosafe,wischnewski2019model,konig2023safe,weber2021safe, holzapfel2024event}. However, it is not always clear how to determine these hyperparameters for a truly unknown objective.

In contrast to SafeOpt, Lipschitz only safe BO (LoSBO) \cite{fiedler2024on} relies on the assumption of an known upper bound of the Lipschitz constant of the unknown function. In addition, bounded noise is assumed. In practice, determining an upper bound of the Lipschitz constant is not trivial. One approach for practitioners could be a sensitivity analysis or using a simulation for the estimation.

Safe BO, especially SafeOpt and its variants, were used in numerous control applications such as quadrotors \cite{berkenkamp2016safe, berkenkamp2023bayesian,holzapfel2024event}, 
rotational drive systems \cite{konig2021safe}, a Furuta pendulum \cite{baumann2021gosafe}, and legged locomotion task  \cite{widmer2023tuning}. \cite{menn2024lipschitz} applied the Lipschitz safe algorithm %
to automotive lateral controller tuning.

\subsection{Crash Constraints} %
\label{sec:p2crash} 

\begin{tcolorbox}[methodbox]
\textbf{Goal:} Optimize while experiments may fail depending on the parameterizations %

\medskip
\textbf{Example:} The system becomes unstable and the experiment has to be stopped prematurely

\medskip
\textbf{Available information:}
\begin{itemize}[left=10pt, itemsep=-3pt, topsep=1pt]
\item Boolean success indicator $l^\mathrm{c}_k$ %
\item Noisy sample of objective $J_{k}$ only available if experiment is successful \hbox{$l^\mathrm{c}_k = 1$} 
\end{itemize}
\end{tcolorbox}

\fakepar{Problem Statement} In some controller tuning problems, (useful) query results $\obj_\boiter$ (and $\boldsymbol{g}_\boiter$) may not be attainable for some parameter choices $\params_\boiter$. For example, imagine querying the cart-pole controller with an unstable parametrization. %
Then, the experiment may have to be stopped to prevent damage to the hardware, or the objective function value would grow extremely large.
Other reasons for episodes not being completed successfully include: 
    operators or automatic safety functions aborting experiments if dangerous system behavior is perceived.

This setting is known as  \textit{learning with crash constraints} (LCC) \cite{marco2021robot}. Crash constraints can be modeled by introducing an additional binary response $l^\mathrm{c}(\params)$. In this case the black box returns an additional binary $l^{\mathrm{c}}_k$ when queried. 
It equals one ($l^{\mathrm{c}}_k = 1$) if the black-box evaluation was successful, and equals zero ($l^{\mathrm{c}}_k = 0$) otherwise. In the case of $l^{\mathrm{c}}_k = 0$, the objective function value is unavailable $J_k = \emptyset$ and cannot be added as a new sample point to the data set. The modified version of the single-objective problem 
can be written as: 
\begin{equation}
	\begin{aligned} \label{eq:ChallOptCrash} 
		\params^* =  \argmin{\params \in \Theta} \qquad &  \mathbb{E}\left[ J(\theta) \right] \\
		& \mathrm{P}(l^\mathrm{c}(\params) = 1) > \alpha.\\
	\end{aligned}
\end{equation} 

The main challenge in the LCC setting is to steer optimization away from the crash region without the availability of the function value.

\fakepar{Methods \& Applications}
A simple heuristic to deal with crash constraints is  a fixed penalty for crashed evaluations (\eg \cite{marco2016automatic,shahrokhshahi2022sample}), or to use data obtained before the crash (\eg \cite{calandra2016bayesian}). However, it may require substantial expert domain knowledge
to design the objective function such that smoothness at the borders between crashed and successful evaluations is preserved. Smoothness at the borders between crashed and successful evaluations is important for using GPR models as surrogate models. 

Another heuristic approach is to introduce virtual data points based on pessimistic GP predictions (BO-VDP) \cite{stenger2022benchmark}. This can be easily combined with other BO settings such as constrained BO \cite{stenger2022joint}, multi-objective BO \cite{stenger2022what}, contextual BO \cite{stenger2023automatic}, and local BO \cite{vonrohr2024local} without adjusting the acquisition function.   

Other approaches include a probabilistic classifier to identify infeasible parameters (\eg \cite{lindberg2015optimization, kato2017multiobjective,chakrabarty2022extremum}). However, this requires training an additional probabilistic classification model with additional hyperparameters.
Alternatives are \citet{marco2021robot,wang2024constrained}. Both introduce GP models capable of combining regression and classification, and combine them with constrained BO. These methods require additional GP models and adjusting the acquisition functions in case of otherwise unconstrained optimization problems.

In some cases, unsuccessful evaluations, \eg extremely large objective evaluations, need to be detected first. For that purpose, for example, \citet{martinez2018practical} proposed a method based on robust regression using a student-t likelihood. %

\subsection{Multi-Objective BO} \label{sec:p2MultObj}

\begin{tcolorbox}[methodbox]
\textbf{Goal:} Simultaneously optimize (potentially) conflicting objectives 

\medskip
\textbf{Example:} Tracking accuracy vs. energy consumption

\medskip
\textbf{Available information:}
\begin{itemize}[left=10pt, itemsep=-3pt, topsep=1pt]
\item Noisy samples of $M$ objectives $(J_{k,1},\dots,J_{k,M})$
\end{itemize}
\end{tcolorbox}

\fakepar{Problem Statement}
There is often more than one objective in control engineering tuning tasks.
Typical examples are the trade-off between energy consumption and tracking performance or rise time and overshoot. 
Conflicting objectives can be combined into one objective by weighted summation.
Alternatively, the problem can be reformulated as a constrained optimization problem (cf. Sec. \ref{sec:p2Constrnt}).
Both methods require additional prior information.
Either the objective function weights have to be specified a-priori or constraint thresholds need to be known. 
In contrast, the goal in multi-objective Bayesian optimization (MOBO) is to simultaneously optimize $M$ (possibly) conflicting objectives $\obj_1(\params), \obj_2(\params), \dots , \obj_M(\params)$: 

\begin{equation}
		 \min_{\params \in \paramSet} \;  \mathbb{E} \left[\obj_1(\params)\right],  \mathbb{E} \left[\obj_2(\params)\right], \dots ,  \mathbb{E} \left[\obj_M(\params) \right]
         \label{eq:ChallOptMult} 
\end{equation}
Each experiment returns a noisy sample of each of the $M$ objectives $(J_{k,1},\dots,J_{k,M})$. Instead of searching for one optimal parametrization $\params^\ast$, the solution to \eqref{eq:ChallOptMult} is a set of non-dominated or Pareto-optimal parameterizations. 
Loosely speaking, a parametrization is Pareto optimal if there is no other parametrization that is better in all performance metrics.
A practitioner can, after the optimization, decide on one parameterization from the Pareto-optimal solution set to balance the conflict between the competing objectives individually. %
The main challenge of MOBO is to propose parameters that improve the current set of Pareto optimal solutions instead of only improving a single metric. %

\fakepar{Methods \& Applications}
It is common MOBO practice to model each objective function with a separate GP model. The acquisition function now represents the utility of improving over the current set of Pareto optimal designs. 

One popular indicator for the quality of a Pareto front is the hypervolume indicator. The expected improvement of the hypervolume (EIHV) \cite{emmerich2008compuation} seeks to increase the hypervolume and is a common MOBO acquisition function. Examples in control and robot learning on real hardware include gait optimization \cite{tesch2013expensive, ariizumi2017mutiobjective}, mobile robotics \cite{kato2017multiobjective}, and trading of performance and robustness \cite{turchetta2020robust}.    

However, more recent work in other domains has shown that Thompson sampling efficient multi-objective optimization (TSEMO) \cite{bradford2018efficient} and the expected improvement-matrix criterion \cite{zhan2017expected} (EIM) perform competitively to the EIHV criterion, while reducing computational overhead. TSEMO was also applied to multi-objective MPC tuning with crash constraints in simulation \cite{stenger2022what}. EIM was used  by \citet{tang2022borderline} for engine calibration on real hardware.

\subsection{Preferential BO} \label{sec:p2preferential} %

\begin{tcolorbox}[methodbox]
\textbf{Goal:} Optimize with (subjective) human feedback given as pairwise comparisons

\medskip
\textbf{Example:} The step response for experiment with $\params_k$ looks better than with  $\params_k^\prime$  

\medskip
\textbf{Available information:}
\begin{itemize}[left=10pt, itemsep=0pt, topsep=1pt]
\item Subjective preferences $p_k$ ($\params_k$ is preferred over~$\params_k^\prime$)
\end{itemize}
\end{tcolorbox}

\fakepar{Problem Statement}
Above, we assumed that the objective can be quantified in mathematical terms and recovered numerically from the trajectory data measured during an episode \eqref{eq:single_objective}. This is, for example, the case when optimizing energy consumption or tracking performance. However, finding an exact mathematical cost function formulation can be challenging when considering more complex requirements such as comfort. This may result in tedious tuning of the cost function, which is often time-consuming and requires multiple optimization runs. Consequently, in some applications, the tuning effort is only shifted from controller parameter tuning to objective function tuning.%

This observation gives rise to BO paradigms where subjective human preferences are used as feedback during optimization. In the narrow sense, preferential BO \cite{gonzales2017preferential} refers to noisy binary feedback: We still want to solve the global optimization problem of finding: 
\begin{equation}
		\boldsymbol{\theta}^* =  \argmin{\boldsymbol{\theta} \in \Theta} \;  \mathbb{E}\left[ J(\theta) \right]
        	 \label{eq:ChallOptPref} 
\end{equation}
However now, we do not receive samples of the objective function $J_k$. Instead, information about $ J(\theta)$ can only be obtained by evaluating in pairs of points or duels $\left[ \params_k, \params_k^\prime \right] \in \Theta \times \Theta$ from which we obtain binary feedback $p_k \in \left\{ 0, 1 \right\}$ that represents whether or not $\params_k$ is preferred over $\params_k^\prime$. %

In the related coactive feedback framework \cite{shivaswamy2012online} the user actively chooses a query instead of the query suggested by the algorithm. This way prior engineering expertise can be included in the tuning process. 
The combination of preferential and coactive feedback is referred to as mixed-initiative learning \cite{lester1999lifelike,tucker2022polar}.   

The main challenge of PBO is that the objective function is not directly accessible, and the acquisition function has to select pairs of parameter combinations for the user to compare.

\fakepar{Methods \& Applications}
Methods addressing the preferential BO setting have been developed mainly outside of the control and robotics community by, \eg \citet{astudillo2023qEUBO,Xu2024Principled, gonzales2017preferential}. The main idea centers around modeling the preference function using a GP $J(\theta)$ and capturing the user preferences through \eg a probit likelihood. Exact posterior GP inference becomes intractable due to the non-Gaussian likelihood, this is why approximations such as the Laplace approximation have to be used.  

Early hardware applications of PBO in robotics are presented by \cite{thatte2017sample,gras2018gaze}. %
A relatively recent PBO framework %
is POLAR \cite{tucker2022polar} that was applied to various robotics applications \cite{cosner2022safety, tucker2020preference, tucker2020human,tucker2021preference,li2022natural,csomay2022learning}. POLAR uses a Thompson sampling approach on subspaces. However, Thompson sampling was reportedly outperformed by more recent PBO variants \eg \cite{astudillo2023qEUBO}. EUBO \cite{astudillo2023qEUBO} has been used for controller tuning in simulation and was shown to satisfy user preferences quicker than random search and multi-objective BO \cite{coutinho2024human}. %

An alternative popular approach for preferential black-box optimization based on a radial-basis function surrogate model is GLISP \cite{bemporad2019active}. 
It was for example applied in \cite{zhu2021cglisp} to an MPC controller for lane-keeping and obstacle-avoidance in autonomous driving and various robotics applications \eg \cite{campagna2024promoting, roveda2023human}. However, to our knowledge, GLISP has not been compared to recent PBO variants.

Subjective human feedback has also been used in BO for robotics and control in hardware experiments (\eg \cite{junge2020improving, matsubara2016data, tran2025mixed, deneault2025preferential}) using real-valued feedback and non-preferential standard BO. However, rating the performance on a real-valued scale is expected to be a lot harder for humans than just deciding whether performance $\params_k$ is superior to performance $\params_k^\prime$.

\subsection{Contextual BO} \label{sec:p2contextualbo}

\begin{tcolorbox}[methodbox]
\textbf{Goal:} Find optimal parameters as a function of a context 

\medskip
\textbf{Example:} 
Gait parameters as a function of surface roughness 

\medskip
\textbf{Available information:}
\begin{itemize}[left=10pt, itemsep=0pt, topsep=1pt]
\item Noisy sample of objective $J_{k}$ given the context~$\context_k$
\end{itemize}
\end{tcolorbox}

\fakepar{Problem Statement}
The choice of optimal controller parametrizations may depend on additional operating conditions. For example, the optimal parametrization of a controller for climate control may depend on the ambient temperature \cite{stenger2023vehicle, fiducioso2019safe}. If performance-critical operating conditions are known before the start of the episode, they can be explicitly considered in the surrogate model in BO to solve the context-dependent optimization problem. From a control engineering perspective, the context can be seen as a disturbance with an impact on controller performance that is sufficiently well represented by a constant over the experimental episode. 

Instead of searching for one optimal parametrization $\params^*$, contextual BO searches for a function $\params^* (\context)$ that gives optimal parameters depending on the context value(s) $\context$, \eg ambient temperature. This function is closely related to gain scheduling from control engineering. 
The contextual unconstrained single-objective problem is stated as follows:  
\begin{equation}
	\begin{aligned} \label{eq:ChallOptContext} 
		\params^*(\context) =  \argmin{\params \in \Theta} \qquad &  \mathbb{E} \left[\obj(\params , \context) \right] \\
		\mathrm{s.t. }\qquad \qquad \qquad \qquad  
		& \boldsymbol{s}_{\mathrm{min}} \le \boldsymbol{s} \le  \mathbf{s}_{\mathrm{max}} 
	\end{aligned}
\end{equation}
In the temperature example, the context is set by the environment. In this case, there are two interactions with the environment. First, the context $\context_\boiter$ is revealed by the environment, and then the BO algorithm suggests parameters for this context by optimizing the acquisition function for that context. They are then evaluated on the black-box function. %

In offline contextual BO, see \cite{char2019offlline,chung2020offline}, the context is also chosen by the BO algorithm. With few exceptions, \eg \cite{le2024controller}, this has received very limited attention in BO for control and robot learning,
A related BO paradigm is multi-task BO \cite{swersky2013multi}, see Sec.~\ref{sec:p2othervariants}.

\fakepar{Methods \& Applications}\label{sec:CEContextLit}
Contextual BO can be realized by adding the contextual variables as additional inputs to the GP of the objective function (and the GP of the constraints). Here, often a product kernel is used. Then, the acquisition function can be maximized for the context provided by the environment. This way, the classical GP-UCB algorithm can be extended to the contextual case \cite{krause2011contextual}. However, care needs to be taken when using expected improvement in the contextual setting because there does not exist a single \emph{best solution}. 

Experimental applications were presented in \cite{frohlich2021model} for an autonomous race car, in \cite{berkenkamp2023bayesian} for a quad-rotor in \cite{konig2021safe} for a rotational motion system, in \cite[Sec. 6.3]{stenger2023automatic} for a three-tank system, in \cite{yu2022learning} for a legged robot, in \cite{zhang2024online} for quadruped locomotion, and in \cite{widmer2023tuning}. 

The contributions \cite{berkenkamp2023bayesian,konig2021safe, widmer2023tuning} demonstrate, how contextual BO can be combined with safe exploration. This combination can be achieved relatively straightforwardly by calculating the safe set only for the current context.

\subsection{Time-Varying BO} \label{sec:p2timevarying}

\begin{tcolorbox}[methodbox]
\textbf{Goal:} Track optimal solution through time

\medskip
\textbf{Example:} Adjust parameters as actuators experience wear

\medskip
\textbf{Available information:}
\begin{itemize}[left=10pt, itemsep=0pt, topsep=1pt]
\item Noisy sample of objective $J_{k}$ at time step \hbox{$\boiter \in \mathbb{I}_{1:T} \coloneqq \{1, \dots, T\}$}
\end{itemize}
\end{tcolorbox}

\fakepar{Problem Statement}
So far we have discussed different approaches to find the optimum of a \emph{time-invariant} objective function.
However, especially in the context of control, the objective may be \emph{time-varying} due to ongoing changes on the dynamical system or changes in the objective.

Considering time-variations in the objective fundamentally changes the optimization problem.
It is no longer of interest to find one optimal solution (or a Pareto front), but the goal is to track an optimal solution through time as
\begin{equation}
    \params_\boiter^* = \arg\min_{\params \in \paramSet} J_\boiter(\params),
\end{equation}
where subscript $\boiter$ denotes the time dependency of the objective and its optimizer.
In this setting, an algorithm chooses a query $\params_\boiter \in \paramSet$ at each time step $\boiter \in \mathbb{I}_{1:T} \coloneqq \{1, \dots, T\}$.
The objective function's time dependency also changes the exploration-exploitation trade-off, as the optimal decision and the information a decision maker gains from each query are now affected by time, and any collected dataset can become stale.
Time-varying BO can also be considered as contextual BO (\cf Sec.~\ref{sec:p2contextualbo}) but with the key characteristic that the time context is strictly increasing and each context only occurs once.

\fakepar{Methods \& Applications}
Finding a suitable strategy to deal with stale data is the core question surrounding research in time-varying BO.
Here, two main strategies have emerged.
The first strategy is to explicitly incorporate assumptions about the temporal changes in the underlying GP model \cite{nyikosa2018bayesian,bogunovic2016time,gao2022bayesian,brunzema2022controller,deng2022weighted,bardou2024too, cho2024run} resulting in a so-called spatio-temporal GP.
The second strategy is to use a time-invariant GP model and cope with stale data by deleting it through resets \cite{bogunovic2016time,zhou2021no,brunzema2022event,holzapfel2024event} or a sliding window~\cite{zhou2021no}.

Both approaches have specific advantages and drawbacks.
When prior knowledge about the temporal dynamics of the objective is available, integrating this knowledge into the GP model can yield superior performance as the surrogate can effectively guide the search. 
However, prior information on temporal properties is not always available or reliable.
In these cases, adaptive methods that detect and respond to changes online show promise \cite{brunzema2022controller,holzapfel2024event}.
These methods face the challenge of fitting a time-invariant GP to time-varying data, which requires careful selection of hyperpriors for parameters like length scales and noise variance, and thus relies more heavily on prior knowledge of spatial properties of the objective \cite{brunzema2022event,bardou2024too}.

Most work in time-varying BO has focused on deriving theoretical guarantees and identifying the regularity assumptions on temporal changes needed to achieve sub-linear regret.
Still, some papers also highlight the practical applications of these approaches, demonstrating time-varying BO yields improved performance in real-world settings.
In the following, we list some of its applications to controller tuning.
For example, event-triggered adaptation in combination with ideas from safe BO (\cf Sec.~\ref{sec:p2safeBO}) has been effectively applied to tuning the position controller of a quadcopter with time-varying dynamics \cite{holzapfel2024event}. 
Similarly, \citet{konig2021safe} combine safety with time-varying BO but using the spatio-temporal model of \citet{bogunovic2016time} instead of an adaptive approach.
They demonstrate the efficacy of their method in the simulation of a controller tuning task of a rotational axis drive under a linear drift of the rotation damping coefficient.
The spatio-temporal model from \citet{brunzema2022controller} was also extended with positional encoding by \citet{cho2024run} for tuning an MPC for a plasma-assisted deposition process.

\subsection{Decision Spaces: Optimize Integer or Categorical Variables} \label{sec:p2decisionspaces}

\begin{tcolorbox}[methodbox]
\textbf{Goal:} Optimize over non-real-valued parameter spaces $\paramSet \not\subseteq \mathbb{R}^d$

\medskip
\textbf{Example:} the prediction horizon of an MPC  

\medskip
\textbf{Available information:}
\begin{itemize}[left=10pt, itemsep=0pt, topsep=1pt]
\item Noisy sample of objective $J_{k}$
\end{itemize}
\end{tcolorbox}

\fakepar{Problem Statement}
An additional challenge for automatic controller tuning are parameter spaces that are not sets of real numbers. The problem is 
\begin{equation}
    \params^* = \arg\min_{\params \in \paramSet} J(\params)
\end{equation}
where $\paramSet \not\subseteq \mathbb{R}^d$.
Examples are integer variables such as the horizon length in MPC or discrete choices such as the usage of an additional filter or feed-forward controller component.

\fakepar{Methods \& Applications}
From a method perspective BO over discrete spaces has two additional challenges. First, we need to find a suitable GP to model correlations in combinatorial spaces and, second, the optimization of the acquisition itself now involves the solution to a combinatorial problem. 

Methods like SMAC \cite{hutter2011sequential} and BOCS \cite{baptista2018bayesian} introduced tree-based surrogates and sparse regression with one-hot encoding, while \cite{dadkhahi2022fourier} improved encoding efficiency. Neural network surrogates \citep{hase2018phoenics, hase2021gryffin}, modified kernels in Casmopolitan \citep{wan2021think}, and discrete diffusion kernels in COMBO \cite{oh2019combinatorial} and HyBO \cite{deshwal2021hybrid} enhanced interaction modeling. To tackle high-dimensional spaces, \cite{deshwal2023bayesian} proposed low-dimensional embeddings for discrete search spaces. The problem of optimizing over discrete spaces can be circumvented by rephrasing the problem into a probabilistic one by optimizing over the continous parameters of discrete probability distributions \cite{daulton2022bayesian}.
Recently, CBOSS \cite{rath2024discovering} has been introduced as method for combinatorial BO with both constraints (Sec.~\ref{sec:p2Constrnt}) and crash constraint (Sec.~\ref{sec:p2crash}).
CBOSS has been applied to find a real-time capable parametrization for a digital twin of an electric vehicle for a driving simulator.
Other studies have applied BO to tune the horizon length in MPC, an integer valued parameter, however, they do not report on the details of their method \cite{lucchini2020torque,stenger2020robust,tao2024goal}.

\subsection{Local BO }  \label{sec:p2high_dim}

\begin{tcolorbox}[methodbox]
\textbf{Goal:} Search for a local optimum that is close to an initial design $\params_0$

\medskip
\textbf{Example:} Incremental improvement over a parametrization derived from a linearized dynamics model

\medskip
\textbf{Available information:}
\begin{itemize}[left=10pt, itemsep=0pt, topsep=1pt]
\item Noisy sample of objective $J_{k}$
\end{itemize}
\end{tcolorbox}

\fakepar{Problem Statement}
While it is generally desirable to find a globally optimal controller parameterization this goal may not be achievable or cost effective in practice.
Whenever the parameter set $\paramSet$ is too large to effectively search through or we already have a good guess where in the domain the optimal parameters are we can resort to \emph{local} optimization. 
Local optimization methods require the user to provide an additional parameter, the initial guess $\params_0$ that implicitly defines the local neighborhood $\paramSet(\params_0)$ within which we restrict the search. The problem formulation is then given as 
\begin{equation}
    \params^* = \arg\min_{\params \in \paramSet(\params_0)} J(\params).
\end{equation}

Local optimization produces a sequence of parameters $\{\params_0, \params_1, \dots\}$ where each set of parameters improves over the previous one and is ``close'' to the previous one. Next to their higher effectiveness in large domains these local search variants can have additional benefits for controller tuning. Local methods exhibit continuous improvement, and local exploration which can avoid evaluating unstable parameters. As an additional benefit, these algorithms can be perceived as more intuitive due to smaller and local updates, making these algorithms easier to deploy and tune. 
Local optimization can also help if a suitable domain of the tuning parameters $\paramSet$ is unknown. Often, these need to be carefully chosen (see for example \cite{roveda2020twostage}) in order for BO to find good solutions in the desired number of iterations.
Nevertheless, local optimization is sensitive to the initial parameters $\params_0$ and may yield sub-optimal solutions. 

\fakepar{Methods \& Applications}
In the literature two main approaches are commonly used, \emph{trust-regions} \cite{park2016bayesian, frohlich2019bayesian, park2020contextual, eriksson2021scalable, frohlich2021cautious} and line-search \cite{kirschner2019adaptive, muller2021local, nguyen2022local, wu2023behavior, brunzema2025bayesqp,tang2025nest}. 
Trust-region BO such as TurBO \cite{eriksson2021scalable} maintain a small subset of the domain $\paramSet$ in which it executes the BO procedure. In TurBO the subset expands when the optimizer is able to improve upon its previous candidate and shrinks when it fails to do so.
Line search methods maintain a current candidate parametrization and iteratively choose a search direction and step size to improve the candidate. For example, gradient information BO (GIBO) \cite{muller2021local} leverages the GP model of the objectives gradient $\nabla J$ to find its search direction and step size.
Similarly, DescentLineBO \cite{kirschner2019adaptive} selects search direction based on a gradient estimation and then a full BO to find the optimum on the subspace defined by the search direction.
BayeSQP~\citep{brunzema2025bayesqp} leverages similar ideas in combination with classic sequential quadratic programming for local constrained BO.
\citet{stenger2026local} extends the information-theoretic entropy search paradigm to local BO.
\citet{menn2026local} investigates several of the local BO approaches discussed above in the context of preferential BO.
In \citet{vonrohr2024local}, the authors use an extended version of GIBO with crash constraints to tune PID, LQR, and MPC controller on a simulated three-tank system and PID controllers on hardware.   
Trust-region based methods are used in the studies by \citet{park2016bayesian} and \citet{park2020contextual} and applied to a wind farm control problem with hardware experiments in \cite{park2016bayesian}.

\subsection{Further Variants } \label{sec:p2othervariants}

We briefly review further BO variants relevant for controller tuning and robotics. 
In \emph{multi-fidelity} optimization, \eg \cite{marco2017virtual, kandasamy2019multi, sorourifar2023computationally, nobar2024guided, he2025simulation, tan2024optimal,tang2024cages}, the goal is to combine information obtained from multiple sources with different accuracy and evaluation costs. In the context of tuning for robotics and control, a simulation may be a low-fidelity information source: it is fast to evaluate but may lack accuracy. In contrast, the real hardware experiment is a high-fidelity information source: it is expensive to evaluate but is highly accurate as it represents the target system. 

In \emph{multi-task} \cite{swersky2013multi} BO knowledge is transferred from one task to another during the learning process. 
The term multi-task BO is rarely used in BO for control or robotics. One exception is \cite{luebsen2024towards}, where safety is combined with multi-task BO. However, the paper \cite{luebsen2024towards} could also be classified as multi-fidelity safe BO. Another exception is \cite{sasaki2020bayesian} where results from previous optimization are reused.

The vast majority of BO for controller tuning conduct experiments of fixed episode length. In contrast, \emph{early stopping BO} (ESBO) detects sub-optimality, \eg oscillations, early in the episode and decides when to stop an episode to reduce overall experimentation time. ESBO is a popular framework in ML hyperparameter tuning (see \eg \cite{swersky2014freeze}) and has recently been transferred to the control domain in simulation \cite{hirt2024time} and hardware experiments \cite{stenger2024early}.

\emph{Batch or parallel} BO considers parallel objective function evaluations.  
Batch BO is less relevant for tuning in control and robotics because usually, only one instance of the physical system is available for optimization. 
A simulation example for hardware and control co-design can be found in \cite{baheri2019combined} or for robot cooking with human feedback \cite{junge2020improving}.

Standard BO assumes that there is no cost associated with changing parameters from one iteration to another. However, imagine a process where one parameter is a temperature and it takes a significant amount of time to change the process temperature in between batch runs. 
This cost incurred by changing the parameters is called \emph{path-dependent or switching cost}. 
Local BO versions implicitly prefer small parameter changes and thus produce less path-dependent costs than global BO versions. Furthermore, specific acquisition functions were developed \eg \cite{folch2022snake,liu2023bayesian}.

Recent work has also begun to integrate large language models (LLMs) into BO.
LLAMBO \cite{liu2024llambo} leverages LLMs for warm-starting and candidate generation, while BORA \cite{cisse2025bora} uses LLMs to incorporate contextual and domain knowledge into the optimization process.
For controller tuning, agentic BO \cite{brunzema2026agenticbo} could combine requirements, documentation, code, and guidance by a domain expert to guide and adapt an uncertainty-aware search.
While this direction may improve sample efficiency, important open questions remain regarding safety and reliability, and these methods still require thorough validation in real-world hardware settings.

\section{Review of Practical Applications of BO}\label{sec:p2applications}

This section systematically reviews applications of BO for controller tuning and robot learning focusing on online parameter tuning directly on hardware according the problem setting described in Sec. \ref{sec:p1problemstatement} and Sec. \ref{sec:p2advanced_variants}. 
Sec. \ref{sec:p2reviewmethod} outlines the paper collection method and results are analyzed according to application domains (Sec.~\ref{sec:p2reviewapplication}), optimized policy (Sec.~\ref{sec:p2reviewcontrollers}), and problem formulation (Sec.~\ref{sec:p2reviewproblemformulation}). Other applications of BO in the control and robotics context are briefly reviewed in Sec.~\ref{sec:p2otherusecases}. 

\begin{tcolorbox}[greybox, title={Key Messages}] 
BO application domains are diverse ranging from lab-scale demonstrators to complex industrial processes (Sec.~\ref{sec:p2reviewapplication}). The optimized control policies are equally diverse spanning simple PI controllers via MPC to the latent space of high-dimensional controllers (Sec.~\ref{sec:p2reviewcontrollers}). Most reviewed applications use the vanilla problem formulation presented in Sec.~\ref{sec:p1BOandGPRIntro} and low dimensional problems (Sec. \ref{sec:p2reviewproblemformulation}). There are many BO applications in control and robotics such as planning tasks, offline tuning of state-estimators or robust controller optimization beyond the criteria defined in Sec.~\ref{sec:p2reviewmethod} (Sec.~\ref{sec:p2otherusecases}).
\end{tcolorbox}

\subsection{Methodology} \label{sec:p2reviewmethod}

We consider papers that meet the following criteria:
(i)~controller tuning that is performed online on hardware, \ie each BO query evaluates a controller or policy on a hardware platform (for other use cases of BO in robotics and control, see Sec.~\ref{sec:p2otherusecases}), and
(ii)~published in 2025 or earlier.

We search all papers with the term `Bayesian optimization' in the title or abstract published in the ten most impactful venues in robotics and automatic control according to Google Scholar %
and the two most prominent control conferences (see Table~\ref{tab:venues}).
\begin{table*}[t]
\footnotesize
\centering
\caption{Reviewed venues for the systematic literature search}
\label{tab:venues}
\begin{tabular}{>{}p{3cm}>{}p{12cm}}
\toprule
\textbf{Category} & \textbf{Venues} \\
\midrule
Control & IEEE Trans.\ on Systems, Man, and Cybernetics: Systems; IEEE Trans.\ on Automatic Control; IEEE Trans.\ on Fuzzy Systems; Automatica; IEEE/CAA Journal of Automatica Sinica; ISA Transactions; IEEE Trans.\ on Control Systems Technology; Annual Reviews in Control; Control Engineering Practice; Intl.\ Journal of Robust and Nonlinear Control; Conference on Decision and Control; American Control Conference \\
\midrule
Robotics & IEEE Robotics and Automation Letters; IEEE Intl.\ Conference on Robotics and Automation; Conference on Robot Learning; Science Robotics; Robotics and Computer-Integrated Manufacturing; IEEE Trans.\ on Robotics; IEEE/RSJ Intl.\ Conference on Intelligent Robots and Systems; Robotics: Science and Systems; IEEE Trans.\ on Automation Science and Engineering; IEEE/ASME Trans.\ on Mechatronics \\
\bottomrule
\end{tabular}
\end{table*}
We additionally include papers on hardware applications of BO from other venues without claiming exhaustiveness.
In total, we reviewed 110 papers meeting the above criteria.

\subsection{Application Domains} \label{sec:p2reviewapplication}

Table \ref{tab:hardware_categories} sorts the reviewed papers according to applications. There is a significant amount (N = 26) of applications to lab-scale demonstrators such as an inverted pendulum, or quadrotors. Those are easily accessible in most control and robotics labs and are primarily used to demonstrate that a new methodology, for example safe exploration 
\cite{berkenkamp2023bayesian} or time-varying BO \cite{holzapfel2024event}, works on hardware in principle. 

Additionally, Table \ref{tab:hardware_categories} reveals cases where BO demonstrates practical benefits beyond lab-scale demonstrators. 
For example \citet{savaia2021experimental,menn2024lipschitz} show that BO allows to automatically calibrate vehicle suspension and path tracking controllers reducing required time and expertise when adapting controller parameters to new vehicles. 

BO was also shown to be particularly useful when experiments are conducted with the human in the loop \cite{li2023hierarchical, kim2019bayesian, wen2020personalization, zahedi2022user,li2022engagement}. Here, controller parameters have to be adapted quickly to best cater to different human needs. When humans are in the loop it is particularly important to make experimentation time as short as possible, \ie have high sample efficiency. 

Other use cases where accurate first principle modeling is difficult and experimentation is expensive are wind farm control \cite{park2016bayesian}, complex industrial processes such as CNC-machines \cite{khosravi2023safety}, and particle accelerators \cite{shalloo2020automation}. Here BO reveals its potential to contribute to sustainable energy production, increased production efficiency, and scientific discovery in other research areas.

\subsection{Optimized Controller Type} \label{sec:p2reviewcontrollers}
Our review shows that BO is indeed very flexible with respect to the optimized policies as postulated in Sec. \ref{sec:p1problemstatement}. Examples include model predictive control (MPC) (\eg \cite{frohlich2021model, widmer2023tuning}), LQR (\eg \cite{marco2017virtual, marco2016automatic}), PID (\eg \cite{roveda2020twostage, khosravi2023safety}), impedance control \cite{chen2023bayesian}, neural network controllers \cite{frohlich2021cautious},  grasping parameters (\eg \cite{nishimura2017thin}, \cite{lechuz2024bayesian}) or gait parameters (\eg \cite{calandra2016bayesian, ariizumi2017mutiobjective,tesch2013expensive}). 
It was also shown that BO can be applied to complex hierarchical 
controller structures \cite{wang2025safe,khosravi2023safety, gabler2022bayesian,roveda2020twostage} and the latent space parameters of high-dimensional policies, \eg \cite{delgado2020contextual, yu2019sim,delgado2020sample,antonova2020bayesian}.
However, Fig. \ref{fig:statistics} shows that the vast majority of hardware examples of BO optimize less than $10$ parameters. To our knowledge, the highest dimensional hardware use case of BO has been reported by \citet{frohlich2021cautious}. This is in contrast to the more than $100$ dimensional problems that are typically addressed in the literature on high-dimensional BO, \eg \cite{hvarfner2024vanilla} in simulation.

\subsection{Problem Formulations} \label{sec:p2reviewproblemformulation}

We have presented in Sec. \ref{sec:p2advanced_variants} that BO can be extended in many ways to address challenges arising in practical controller tuning tasks.
Table \ref{tab:hardware_extensions} shows that real-world efficiency gains can be achieved even with the vanilla BO formulation (N = 57).
The most popular extensions are unknown constraints, safe exploration and contextual BO. Still Table \ref{tab:hardware_extensions} shows that the current state of the art of BO methods is further developed than what is used in real-world applications beyond lab-scale demonstrators.  

Additionally, Table \ref{tab:hardware_extensions} reveals that if advanced problem formulations are used, most papers consider a single extension. Exceptions are for example safe-contextual \cite{konig2021safe}, safe-time-varying \cite{holzapfel2024event} or local-crash-constrained BO \cite{vonrohr2024local}.

\begin{table*}[ht] 
	\caption{BO paper with hardware experiments sorted with respect to application categories } \label{tab:hardware_categories}
	\centering
	\footnotesize
	\begin{tabular}{ >{\raggedright}p{3cm}  >{\raggedright}p{12cm}  }
		\toprule
		Category & References
		  
 \tabularnewline 
 \midrule 
Lab-scale demonstrators (${N = 26}$) & Cart Pole \citet{marco2017virtual}, Pole Balancing with Robot Arm \citet{marco2016automatic,he2025simulation}, Quadrotor \citet{duivenvoorden2017constrained,berkenkamp2023bayesian,bensal2017goal,berkenkamp2016safe,holzapfel2024event}, Rotational Motion System \citet{konig2021safe}, Servo Motor \citet{chen2019controller}, Furuta Pendulum \citet{frohlich2021cautious,turchetta2020robust,frohlich2019bayesian,baumann2021gosafe}, Three Tank System \citet{vonrohr2024local,stenger2023automatic,stenger2024early}, Linear Motion System \citet{khosravi2022performance}, Ball Throwing \citet{yamawaki2024motion}, Miniature Autonomous Racing \citet{puigjaner2025performance}, Rotary Motor System, Linear Motor System \citet{nobar2024guided}, RC Race Car \citet{frohlich2022contextual}, Temperature Control on Embedded Platform \citet{bakaravc2025bayesian}, Mini Wheelbot \citet{hose2025mini}, Turtlebot 3 \citet{hossen2025cure}, Mini Wheelbot, Cartpole \citet{hose2024fine} 
 \tabularnewline 
 \midrule 
Locomotion (${N = 28}$) & Quadruped Robot \citet{marco2021robot,zhu2019bayesian,zhai2022scaffolded,zhang2024online,cosner2022safety}, Bipedal Walker Fox \citet{calandra2016bayesian,calandra2014bayesian}, Snake Robot \citet{ariizumi2017mutiobjective,tesch2013expensive,tesch2011using}, Tracked Mobile Robot \citet{kato2017multiobjective}, Legged Robot \citet{widmer2023tuning}, 6-Legged Robot \citet{pautrat2018bayesian}, Laser-Plasma Accelerator \citet{jalas2021bayesian}, Bipedal Locomotion \citet{yang2022bayesian}, Gait Optimization \citet{lizotte2007automatic}, Humanoid Robot \citet{hemker2009efficient}, Legged Robot Climbing \citet{yu2022learning}, Soft Quadruped Robot \citet{tan2024optimal}, Mini Rover turning \citet{kerimoglu2024learning}, Continuous Quadruped Jumping \citet{bellegarda2024quadruped}, Micro Robot \citet{vonrohr2018gait}, Robot Locomotion \citet{yu2019sim}, Robot Hopping \citet{galljamov2021adjustable}, Hexapod, Manipulation \citet{antonova2020bayesian}, ATRIAS Robot \citet{antonova2017deep}, Magneto-Elastomer Robot Walking \citet{culha2020learning}, Robot Balancing \citet{kuindersma2012variational} 
 \tabularnewline 
 \midrule 
Industrial and Process Control Systems (${N = 11}$) & Throttle Valve Control \citet{neumannbrosig2019data}, CNC Machine \citet{khosravi2023safety}, Laser Wakefield Accelerators \citet{shalloo2020automation}, Laser Powder Bed Fusion \citet{kavas2024situ}, Laser-Plasma Accelerator \citet{jalas2021bayesian}, Free Electron Laser \citet{kirschner2019adaptive}, Precision Motion System \citet{konig2023safe}, Current and Voltage Control \citet{weber2021safe}, High-Precision Motion Systems in Semiconductor Industry \citet{konig2025adaptive}, Waste Crane \citet{sasaki2020bayesian}, Permanent Magnet Synchronous Motor \citet{wang2025safe} 
 \tabularnewline 
 \midrule 
Robotics Manipulation (${N = 17}$) & 7-DOF Robot Arm \citet{roveda2020twostage}, Robotic Screw Insertion \citet{gabler2022bayesian}, Robot Manipulation \citet{drieß2017contrained}, Robotic Manipulation \citet{chen2023bayesian,rofer2024bayesian}, Robotic Wiping \citet{okada2023learning}, Robotic Trajectory Tracking \citet{roveda2020robot}, Robotic Force Controlled Assembly Processes \citet{cheng2014online}, Reaching Task \citet{guzman2022bayesian}, Soft Robotic Manipulation \citet{zwane2024learning}, Robot Grasping \citet{lechuz2024bayesian,daniel2014active}, Robotic Soft Gripper \citet{nishimura2017thin}, Table Wiping, Rope Winding, Food Stirring \citet{yang2022learning}, Hexapod, Manipulation \citet{antonova2020bayesian}, Robot Pouring \citet{guevara2017adaptable}, Robot Door Opening \citet{englert2016combined} 
 \tabularnewline 
 \midrule 
Automotive (${N = 5}$) & Autonomous Racing \citet{frohlich2021model,wischnewski2019model}, Semi Active Suspensions \citet{savaia2021experimental}, Assisted Gearshift Settings \citet{catenaro2025automatic}, Combustion Engine \citet{tang2022borderline} 
 \tabularnewline 
 \midrule 
Medical (${N = 17}$) & Exosuit \citet{kim2019bayesian,ding2018human}, Spatiotemporal Neurostimulations for Targeted Motor Outputs \citet{laferriere2020hierarchical}, Robot-Assisted Upper Limb Training \citet{zhang2023bayesian}, Transfer Assistant Robot \citet{matsubara2016data}, Robotic Knee Prothesis \citet{li2023hierarchical}, Assist-As-Needed Robot-Aided Therapy \citet{li2022engagement}, Knee Exoskeleton Flexion Assistance \citet{wang2025human}, Surgical System \citet{gras2018gaze}, Back-Support Exoskeleton \citet{sochopoulos2023human}, Plasma Medicine \citet{chan2023towards}, Robotic Feeding \citet{delgado2020sample}, Ankle-Foot Prosthesis \citet{wen2020personalization}, Transfemoral Prosthesis \citet{thatte2017sample}, Exoskeleton \citet{shahrokhshahi2022sample}, Arm Exoskeleton \citet{hamaya2019exploiting}, Feeding Task and Shoe Fitting \citet{delgado2020contextual} 
 \tabularnewline 
 \midrule 
Other (${N = 8}$) & Robotic Cooking \citet{junge2020improving}, Camera Attribute Control \citet{kim2020proactive}, Active Flow Control \citet{blanchard2021bayesian}, Wind Farm Control \citet{park2016bayesian}, Human-Robot Interaction \citet{zahedi2022user}, Sensor Pose Optimization \citet{roveda2021enhancing}, Robot Hockey \citet{kaushik2022safeapt}, Leveling System for Combine Harvesters \citet{dettu2023modeling}\tabularnewline	
\bottomrule	
\end{tabular}
\end{table*}

\begin{table*}[ht] 
		\caption{BO paper with hardware experiments sorted with respect to advanced problem formulations}  \label{tab:hardware_extensions}
	\centering
	\footnotesize
	\begin{tabular}{ >{\raggedright}p{3cm}  >{\raggedright}p{12cm}  }
		\toprule
		Extensions & References 
 \tabularnewline 
 \midrule 
Vanilla BO (No Extensions),  Sec.~\ref{sec:p2vanillaBO} (${N = 57}$) & \citet{marco2016automatic, neumannbrosig2019data, roveda2020twostage, calandra2016bayesian, chen2019controller, savaia2021experimental, frohlich2019bayesian, bensal2017goal, calandra2014bayesian, kim2020proactive, blanchard2021bayesian, shalloo2020automation, khosravi2022performance, laferriere2020hierarchical, zhang2023bayesian, kavas2024situ, pautrat2018bayesian, jalas2021bayesian, matsubara2016data, zahedi2022user, chen2023bayesian, li2022engagement, lizotte2007automatic, tesch2011using, hemker2009efficient, zhu2019bayesian, zhai2022scaffolded, yamawaki2024motion, roveda2020robot, roveda2021enhancing, catenaro2025automatic, wang2025human, cheng2014online, kerimoglu2024learning, bellegarda2024quadruped, guzman2022bayesian, vonrohr2018gait, rofer2024bayesian, zwane2024learning, lechuz2024bayesian, sochopoulos2023human, yu2019sim, delgado2020sample, wen2020personalization, yang2022learning, shahrokhshahi2022sample, galljamov2021adjustable, ding2018human, antonova2020bayesian, antonova2017deep, guevara2017adaptable, daniel2014active, culha2020learning, kuindersma2012variational, hamaya2019exploiting, bakaravc2025bayesian, hose2025mini} 
 \tabularnewline 
 \midrule 
Unknown Constraints,  Sec.~\ref{sec:p2Constrnt} (${N = 10}$) & \citet{marco2021robot, duivenvoorden2017constrained, khosravi2023safety, vonrohr2024local, stenger2023automatic, drieß2017contrained, li2023hierarchical, yu2022learning, dettu2023modeling, englert2016combined} 
 \tabularnewline 
 \midrule 
Safe Exploration,  Sec.~\ref{sec:p2safeBO} (${N = 16}$) & \citet{konig2021safe, berkenkamp2023bayesian, berkenkamp2016safe, widmer2023tuning, baumann2021gosafe, wischnewski2019model, kirschner2019adaptive, konig2023safe, weber2021safe, yang2022bayesian, holzapfel2024event, konig2025adaptive, puigjaner2025performance, nishimura2017thin, kaushik2022safeapt, wang2025safe} 
 \tabularnewline 
 \midrule 
Crash Constraints,  Sec.~\ref{sec:p2crash} (${N = 5}$) & \citet{marco2021robot, kato2017multiobjective, vonrohr2024local, stenger2023automatic, gabler2022bayesian} 
 \tabularnewline 
 \midrule 
Multi-Objective,  Sec.~\ref{sec:p2MultObj} {(${N = 7}$)} & \citet{ariizumi2017mutiobjective, tesch2013expensive, turchetta2020robust, okada2023learning, chan2023towards, tang2022borderline, hossen2025cure} 
 \tabularnewline 
 \midrule 
Preferential,  Sec.~\ref{sec:p2preferential} {(${N = 3}$)} & \citet{gras2018gaze, thatte2017sample, cosner2022safety} 
 \tabularnewline 
 \midrule 
Context,  Sec.~\ref{sec:p2contextualbo}) {(${N = 9}$)} & \citet{frohlich2021model, konig2021safe, berkenkamp2023bayesian, stenger2023automatic, widmer2023tuning, yu2022learning, frohlich2022contextual, zhang2024online, delgado2020contextual} 
 \tabularnewline 
 \midrule 
Time-Varying,  Sec.~\ref{sec:p2timevarying}) {(${N = 2}$)} & \citet{holzapfel2024event, konig2025adaptive} 
 \tabularnewline 
 \midrule 
Local,  Sec. \ref{sec:p2high_dim} {(${N = 5}$)} & \citet{frohlich2021cautious, vonrohr2024local, park2016bayesian, he2025simulation, hose2024fine} 
 \tabularnewline 
 \midrule 
Other Variants,  Sec.~\ref{sec:p2othervariants} {(${N = 8}$)} & \citet{marco2017virtual, kim2019bayesian, junge2020improving, tan2024optimal, nobar2024guided, stenger2024early, he2025simulation, sasaki2020bayesian}\tabularnewline	
\bottomrule	
\end{tabular}
\end{table*}

\subsection{Other Use Cases in Control and Robotics} \label{sec:p2otherusecases}

This section briefly summarizes other use cases of BO in control and robotics that do not match the problem definition in Sec. \ref{sec:p1problemstatement} or the review criteria in Sec. \ref{sec:p2reviewmethod}.

Above we discussed that BO can be used to tune the parameters of control policies.
However, concepts from BO can be used directly for \emph{path planning tasks and environment exploration} \cite{souza2014bayesian, samaniego2021bayesian,shyam2022movement,xu2024online,kiessling2024efficient,garciabarcos2021robust,martinez2009bayesian, marchant2012bayesian}. This is especially appealing in cases of achieving maximum information gain about some uncertain physical quantity \eg bathymetry data \cite{kiessling2024efficient} or tissue abnormalities \cite{salman2018trajectory}. The main difference to the problem formulation in Sec. \ref{sec:p1problemstatement} is the GP modeling an unknown environment quantity instead of the closed-loop control performance.

Similarly, concepts from BO can also be applied to find control inputs in repetitive tasks directly. 
The so-called \emph{run-to-run (R2R) control} paradigm \cite{liu2018survey} is mostly known in  
biomedical engineering and semiconductor manufacturing (EWMA)\cite{liu2018survey}. 
Traditional methods for R2R control are ILC, MPC, and exponentially weighted moving average \cite{liu2018survey}. BO is increasing in popularity for the R2R paradigm \cite{moya2020run,cho2024run}.

Above we discussed how control algorithms can be optimized for an already existing hardware system and a fixed controller.
However, BO can also support control engineers and researchers during controller and hardware (co-)design. For example simulation studies with high-fidelity models can be used to \emph{fairly compare different controller structures} (\eg \cite{manhães2017framework, yu2020comparison,funk2021benchmarking}). 
This way, it can be assessed objectively and automatically which controller structure is most suited for a given task before deployment. 
 
Similarly, the whole \emph{guidance navigation and control (GNC) stack} for AUVs \cite{stenger2022joint}, mobile robots \cite{hossen2025cure} and parafoil landings \cite{polonio2022bayesian} can be jointly optimized with BO in simulation. By using automatic tuning, control engineers can quickly identify whether the GNC structure is feasible and quickly iterate between different structures. 

The optimal controller structure and parametrization depend on hardware design and in turn the hardware sizing depends on how efficiently the controller can use the available hardware resources. Thus, it may be beneficial to simultaneously optimize controller and hardware. Examples for \emph{controller and hardware co-design} have been presented, \eg for airborne Wind Energy Systems \cite{baheri2019combined}, a building chiller plant \cite{bhattacharya2021control}, a robotics system \cite{kim2021mobbo}, a control-on-a-chip system \cite{chan2024practical} and a building energy system \cite{sorourifar2023computationally}. However, most of those works lack experimental validation of the optimized design. One exception is presented by \citet{bjelonic2023learning} where the optimized control and design of a quadruped is evaluated in real-world experiments.

\emph{State estimation, perception and fault/anomaly detection} are sub-fields of control and robotics in which algorithms do not necessarily have to be optimized online. Instead labeled experimental data can be used offline to find the best parameterization. %
This often results in expensive-to-evaluate black-box objective functions where the optimized algorithms are simulated offline with varying parameters on real data. %
Thus numerous applications of automated observer design \cite{chakrabarty2021safe, delcaro2026twin} and Kalman filter tuning with recorded \eg \cite{geht2020environmentally,bertipaglia2022two} or synthetic \eg \cite{nitsch2023automated,nikolaos2019practical,chen24kalman, chen2018weak, chen2019kalman} data have been reported. Optimization of fault diagnosis algorithms has been done  
with experimental data, \eg \cite{stenger2022auto}, and simulation data, \eg \cite{bobrinskoy2012model,marzat2010automatic,marzat2011min,marzat2012robust,marzat2013worst}. Additionally, the optimization of perception parameters dates back to at least \cite{hu2017efficient}.

In some cases, it may be desirable to obtain \emph{robust policy parameters in simulations}. For example, one may want to obtain an initial robust starting point for safe or local online optimization \cite{yang2022bayesian}. 
Due to the model-plant mismatch between simulation and real-world, care needs to be taken not to overfit the policy to simulator dynamics. Instead, the policy shall be robust towards these uncertainties.  

In the easiest case, the optimization can be executed on different randomly generated seeds \cite{sorourifar2021data,stenger2022joint} and a robust constrained formulation can be used (see \eqref{eq:p2unknownconstraints}) \cite{stenger2023vehicle}. More advanced approaches include verifying controllers against adversarial counter examples \cite{ghosh2018verifying,paulson2022adversarially} such as worst-case model-plant mismatch \cite{stenger2020robust,stenger2023automatic}. BO can also be used for data-efficient domain randomization \cite{muratore2021data}.
Alternatively, robustness can be increased by using formulations similar to multi-task BO \cite{toscanopalmerin2015bayesian, groot2010bayesian}, %
robustness towards uncertain input parameters is for example considered by \citet{frohlich2020noisy}, and a multi-objective problem trading off robustness and performance is formulated in \cite{turchetta2020robust}.

\FloatBarrier

\part{Benchmarking}

Systematic benchmarking has been central to progress in machine learning, computer vision, and robotics. It is particularly important for BO, where performance depends strongly on modeling assumptions, acquisition functions, initialization strategies, hyperparameter fitting, and implementation details. In control and robotics, BO methods are often evaluated on individual case studies. While such demonstrations are valuable, they make it difficult to assess whether a method generalizes beyond a specific task or whether its performance depends on favorable design choices.

Part~C discusses how BO methods for control and robotics can be benchmarked more systematically. Section~\ref{p3:howtobenchmark} summarizes recommended problem classes, baselines, metrics, and statistical analyses. A recurring difficulty is the limited availability of realistic, public controller-tuning benchmarks. To make the discussion concrete, we provide a small code template, \textsc{TuneControl} (\href{https://github.com/Data-Science-in-Mechanical-Engineering/tunecontrol}{\faGithub}), which currently contains 34 variants of the popular cart pole and cascaded tank tasks and illustrates how such tasks can be shared in a reproducible format. Section~\ref{sec:p3examplebenchmark} then demonstrates a minimal benchmark comparing common acquisition functions on one of those tasks.

\section{Best Practices in Benchmarking Black-Box Optimization Algorithms} \label{p3:howtobenchmark}

Benchmarking BO methods requires four main choices: (i) benchmark problems, (ii) baseline algorithms, (iii) performance metrics, and (iv) statistical analysis. Algorithm benchmarking is itself an active research area \citep{bartz2020benchmarking, beiranvand2017best,willemsen2024methodology}. We therefore do not attempt to provide a complete treatment, but instead summarize BO-oriented best practices that are particularly relevant for control and robotics.

We focus on vanilla BO, i.e., single-objective unconstrained BO, as introduced in Secs.~\ref{sec:p1BOIntro} and \ref{sec:p2vanillaBO}. The same general principles also apply to the advanced BO variants discussed in Sec.~\ref{sec:p2advanced_variants}, but the metrics and baselines must then be adapted to the respective setting, such as constraints, safety, multiple objectives, contextual variables, or multi-fidelity evaluations.

\begin{tcolorbox}[greybox, title={Key Messages}]
Typical benchmark problem categories range from GP-samples and synthetic test functions to application-oriented simulation environments (Sec.~\ref{sec:p3benchnmark}).  Baselines (Sec.~\ref{sec:p3baselines}) should at minimum include Sobol sampling. Other black-box optimization frameworks are often informative comparators. The most common BO performance metric is regret. However, regret cannot be easily calculated when departing from simple benchmark problems, where the true optimum is known (Sec.~\ref{sec:p3metrics}). Finally, careful statistical analysis is crucial when benchmarking across different seeds and objective functions (Sec.~\ref{sec:p3statstics}). The following checklist serves as practical guideline for reporting benchmarking results:

\begin{checklist}
\item Task, search space and evaluation budget
  \item Number of seeds and repetitions
  \item Initial design and sampling strategy
  \item Surrogate model (kernel, likelihood, hyperpriors)
  \item Acquisition function and all hyperparameters
  \item Baselines and all hyperparameters
  \item Evaluation metric (online curve + final metric)
  \item Overhead/runtime reporting
  \item Statistical test
  \item (for control problems) check state trajectories
\end{checklist}
\end{tcolorbox}

\subsection{Benchmark Problems Categories} \label{sec:p3benchnmark}

Evaluating a BO method on a single objective function is rarely sufficient. The chosen objective may accidentally match the assumptions of one method, for example its kernel, smoothness assumptions, or exploration strategy. Reliable evidence therefore requires benchmark problems that cover different structures, noise levels, dimensionalities, and degrees of model mismatch.

We distinguish three broad categories of benchmark problems: GP-sample objectives, synthetic analytic test functions, and application-oriented simulation or hardware tasks. GP-sample objectives can further be used for within-model or out-of-model comparisons. Across all categories, benchmark results should be reported over multiple random seeds, preferably with shared initial designs across methods. Hyperparameters should be chosen using a procedure that does not leak information from the final test instances into the evaluation.

\fakepar{GP Samples} %
GP samples are functions drawn from the distribution over functions that is defined by a GP prior. They are especially common in the machine learning literature because they make it easy to generate many test functions with controlled properties, such as smoothness, length scales, and observation noise. One practical way to generate analytic GP samples is given by \citet{wilson2020efficiently}.\footnote{Available in \href{https://botorch.readthedocs.io/en/stable/sampling.html\#module-botorch.sampling.pathwise.prior_samplers}{BoTorch}.}

GP samples are useful for studying BO behavior under controlled assumptions. In a \emph{within-model comparison}, the GP prior used by BO matches the GP prior from which the objective functions are sampled. The kernel class, length scales, observation noise, and other prior assumptions are therefore known to the optimizer. This setting represents an idealized case for BO. It is useful for isolating the effect of acquisition functions because it largely removes model mismatch and hyperparameter uncertainty.

In an \emph{out-of-model comparison}, the objective is still sampled from a GP, but the optimizer does not receive the generating hyperparameters. Instead, it must infer them online. The objective may still lie within the assumed model class (\eg correct kernel class), but uncertainty in hyperparameters better reflects practical use.

\fakepar{Synthetic Test Functions}
Synthetic benchmarks provide fast-to-evaluate objectives with known analytic form. A prominent example is the COCO benchmark \cite{hansen2021coco}. Their main advantage is that the global optimum is often known, enabling regret-based evaluation. Moreover, synthetic functions can be chosen to emphasize specific challenges (\eg many local optima in Ackley; narrow valleys in Rosenbrock).\footnote{For an overview of common test functions, see \href{https://www.sfu.ca/~ssurjano/optimization.html}{this collection}.}

Synthetic functions are useful for controlled experiments, debugging, and comparing methods under known difficulty patterns. However, they may not capture the structure of controller-tuning or robotics problems. Therefore, they should not be the only evidence for claims about BO in control and robotics.

\fakepar{Realistic Controller Tuning and Robotics Tasks}
Ap\-pli\-ca\-tion-oriented benchmarks are closest to the settings in which BO is ultimately used. They may be based on hardware experiments, experimentally validated simulations, or simplified but representative simulation environments. Ideally, such benchmarks should specify the controller structure, parameter bounds, objective function, noise sources, failure handling, evaluation budget, and software interface.

Compared with machine learning and AutoML, where many public benchmark suites are available (\eg JAHS-Bench-201 \cite{bansal2022jahs}, HPOBench \cite{eggensperger2021hpobench}), realistic public benchmarks for controller tuning and robotics remain less common \cite{zhou2024impact}. This limits comparability between BO methods and makes it harder to identify robust default settings. To lower the barrier to sharing controller-tuning and robotics benchmarks, we provide \textsc{TuneControl} (\href{https://github.com/Data-Science-in-Mechanical-Engineering/tunecontrol}{\faGithub}): a lightweight benchmark suite for black-box controller tuning.

\begin{tcolorbox}[bluebox, title={TuneControl: A lightweight benchmark for black-box controller tuning. \hfill {\it open~in}~\href{https://github.com/Data-Science-in-Mechanical-Engineering/tunecontrol}{\faGithub}}]

TuneControl provides reproducible controller-tuning tasks with a consistent API.  
It currently includes 34 variants of the popular cart pole and cascaded tank tasks. 
TuneControl is easily extensible, and we invite the community to contribute additional tuning tasks.

\end{tcolorbox}

\subsection{Baselines} \label{sec:p3baselines}
Baselines serve two roles. First, they calibrate the difficulty of the benchmark task. Second, they show whether the modeling overhead of BO is justified. At minimum, BO benchmarks should include Sobol sampling. Sobol sampling is a space-filling non-model-based baseline and is usually stronger than i.i.d.\ random search for low-dimensional bounded domains.

For larger evaluation budgets, it is useful to include low-overhead global optimizers such as CMA-ES \citep{pitra2016doubly}, pattern search, or particle swarm optimization. These methods do not build probabilistic surrogate models and can therefore be competitive when function evaluations are cheap relative to BO overhead. Including such baselines helps determine whether BO is actually needed for the considered budget regime.

Comparing against response surface based optimization can also be informative. Examples include BADS \citep{acerbi2017practical}, GLIS \citep{bemporad2020global}, and SMGO \citep{sabug2021smgo}. These methods are rarely included in BO comparisons, although they may achieve similar sample efficiency in some black-box optimization settings.

Baseline hyperparameters should be treated as carefully as the hyperparameters of the proposed BO method. If the proposed method is tuned for the benchmark, then the baselines should receive comparable tuning effort. Otherwise, the comparison may overstate the advantage of the proposed method. In addition to external baselines, ablation studies are often useful. They compare variants of the proposed algorithm and showcase which components are responsible for performance improvements.

\subsection{Performance Metrics} \label{sec:p3metrics}

The choice of the performance metric depends on what is observable and what is known about the objective. When the global optimum and the latent noise-free objective are available (\eg synthetic functions), regret provides a standard notion of sample efficiency. When they are not available (\eg hardware or realistic simulators), observed-cost and runtime are useful metrics for performance evaluation. 

One common BO performance metric is \emph{simple regret}. It quantifies the sub-optimality of the parameters suggested by an optimization algorithm after a given budget~$k$:
\begin{equation} \label{eq:p3regret}
r_k = \bar{\obj} (\hat{\params}_k) - \bar{\obj}^{*}, 
\end{equation}
where $\hat{\params}_k$ is the recommendation after $k$ evaluations, $\bar{\obj}$ is the latent noise-free objective function evaluated at the location of the estimated best parameter value $\hat{\params}_k$ (see Sec. \ref{sec:p2otherpractical}), and $\bar{\obj}^{*}$ is the global optimum. Simple regret is relevant when the cost incurred during optimization is not important and only the final recommendation matters.

In contrast, \emph{cumulative regret} measures the cost accumulated during optimization; this is the sum of the so-called instantaneous regret:
\begin{equation}
R_k=\sum_{\kappa=1}^{k}\big(\bar J(\theta_\kappa)-\bar J^{*}\big).
\end{equation}
Cumulative regret considers the cost encountered over the whole optimization. Cumulative regret is relevant if parameters are tuned online, \eg in a productive production environment, whereas simple regret is relevant when the cost encountered during optimization is not of interest.

To \emph{practically calculate regret}, the true optimum of the objective function has to be known. This is only the case for synthetic functions that can be optimized globally (see Sec. \ref{sec:p3benchnmark}). For all other functions, the global optimum needs to be estimated in different ways \eg by gridding, or taking the best result achieved by any algorithm on any seed. Additionally, the latent, noise-free objective function $\bar \obj$ has to be known. This is only the case, if the noise is added artificially after calculating $\bar \obj$, for example in GP-samples or synthetic functions.

A more practical alternative to regret is to use the \emph{expected cost} of the optimum parameters at each iteration $\bar \obj(\hat{\params}_\kappa)$. However, this is only possible in benchmarks where $\bar \obj$ is known, \eg in GP-samples or synthetic functions. In particular, it is not known when evaluating on hardware or noisy simulations. In those cases $\bar \obj(\hat{\params}_\kappa)$ may have to be estimated from multiple samples.  

When regret and $\bar \obj(\hat{\params}_\kappa)$ is unavailable, a common fallback is to summarize the observed objective values. However, different summaries have different failure modes: Just reporting the \emph{lowest observed cost} $\min_{\kappa \in \{1,...,k\}} J_{\kappa}$ has to be treated with caution especially in high-noise cases because it is prone to outliers.  

In contrast the \emph{cumulative observed cost}
$\sum_{\kappa=1}^k J_{\kappa}$
is a practical and sound alternative to the cumulative regret as it accurately describes the cost created during optimization.  

Another key metric that should always be included in BO algorithm benchmarking is the \emph{wall clock time} or overhead produced by BO. The computation time is key to evaluate in which circumstances the particular BO algorithm is practical. 

For control and robotics benchmarks, scalar performance metrics alone can be insufficient to verify that an optimized controller behaves as intended. We therefore recommend reporting behavioral diagnostics such as state trajectories alongside the primary optimization metrics. Behavioral diagnostics are not intended to replace the primary scalar objective, but to increase interpretability and ensure that reported improvements correspond to qualitatively reasonable closed-loop behavior.

Extensions of BO require metrics aligned with the extension: When looking \emph{beyond single objective optimization} (Sec. \ref{sec:p2advanced_variants}) metrics specific to each use case have to be used. For example a popular metric for multi-objective optimization is the hypervolume indicator \cite{guerreiro2021hypervolume} with alternatives presented in the overview paper \cite{nery2015performance}. When constraints or safety limits exist, additional metrics such as the number of constraint violations can be informative. 

\subsection{Statistical Analysis}\label{sec:p3statstics}
Meaningful algorithm comparisons require repeated evaluations. At minimum, each method should be evaluated over multiple random seeds. Whenever possible, methods should also be tested on multiple objective functions, task instances, or initial conditions. For fair comparisons, different methods should use the same random seeds and the same initial designs when applicable.

For a fixed task, we recommend reporting the median and interquartile range across seeds. BO performance distributions are often skewed and heavy-tailed, so the mean and standard deviation can be misleading when reported alone. Means and standard deviations may still be useful as supplementary summaries, especially when they are required by a specific reporting convention.

Statistical tests can help assess whether observed differences are likely to be meaningful. Choosing a statistical test is an active research field \cite{demvsar2006statistical,derrac2011practical,carrasco2020recent}, we only suggest one variant. For pairwise comparisons with matched seeds or task instances, the Wilcoxon signed-rank test is a common non-parametric option \citep{wilcoxon1945individual}. It should not be confused with the Wilcoxon rank-sum test, which is used for unpaired samples. For comparisons involving more than two algorithms, a Friedman test can first be used to test whether at least one method differs from the others. If post-hoc pairwise tests are then performed, multiple-comparison corrections such as Holm's procedure should be applied \citep{holm1979simple, garcia2008extension}.

Aggregating results across different objective functions is more difficult because objective values can have different scales. Directly averaging costs or regrets across tasks may therefore be misleading. Common alternatives include ranking methods per task, reporting the fraction of solved target values, or normalizing performance relative to a common baseline \cite{hansen2021coco}.

\section{Example Benchmark}
\label{sec:p3examplebenchmark}

We now illustrate the previous recommendations using a small benchmark on a cart-pole swing-up task. The benchmark compares three common acquisition functions, expected improvement (EI), upper confidence bound (UCB), and max-value entropy search (MES), against Sobol sampling. All BO variants use the same GP surrogate model and differ only in the acquisition function.

The example addresses the following question: under a fixed GP model and without task-specific acquisition-function tuning, how much does the acquisition function affect performance on this controller-tuning task? The goal is not to determine which acquisition function is generally superior. Instead, the example provides a template for reporting a minimal BO benchmark in a truly black-box setting.

\begin{tcolorbox}[greybox, title={Key Messages}]
This section illustrates the benchmarking recommendations from Sec.~\ref{p3:howtobenchmark} by comparing BoTorch implementations of three acquisition functions on a four-dimensional cart-pole controller-tuning task. The example is intentionally small. Its purpose is not to identify the best acquisition function in general, but to demonstrate how a minimal, reproducible BO benchmark can be reported.
\end{tcolorbox}

\subsection{Benchmark Setup}

\begin{table}[!t]
\centering
\captionsetup{width=\linewidth,font=scriptsize}
\caption{Experimental setup, surrogate model, acquisition functions, and baseline.}
\label{tab:p3_benchmark_setup}
\footnotesize
\renewcommand{\arraystretch}{1.0}
\setlength{\tabcolsep}{3.5pt}
\begin{tabularx}{\columnwidth}{@{}>{\raggedright\arraybackslash}p{0.36\columnwidth}>{\raggedright\arraybackslash}X@{}}
\toprule
\rowcolor{myblue!42}
\textbf{Item} & \textbf{Details} \\
\midrule
\rowcolor{myblue!18}
\multicolumn{2}{l}{\textbf{Experimental setup}} \\
Task & Cart-pole swing-up \\
Bounds & $[-3.4,-2.0]\times[-8.0,-4.0]$\newline $\times[-50.0,-30.0]\times[-10.0,-2.5]$ \\
BO bounds & $[0,1]^4$ \\
Budget & $40$ evaluations \\
Seeds & $20$ per method \\
Initial design & Sobol; $5$ initial points \\
\addlinespace[0.3em]
\rowcolor{myblue!18}
\multicolumn{2}{l}{\textbf{Surrogate model}} \\
Kernel & Mat\'ern $\nu=5/2$ \\
Mean & Zero mean \\
Hyperparameters & ARD in 4D \\
Initial lengthscale & $\sqrt{4}$ \\
Lengthscale constraint & $[0.005,4.0]$ \\
Likelihood & Gaussian \\
Fit & MAP \\
Noise constraint & $[10^{-5},10^{-1}]$ \\
Noise fit & MLL \\
\addlinespace[0.3em]
\rowcolor{myblue!18}
\multicolumn{2}{l}{\textbf{Acquisition functions / baseline}} \\
LogEI & No additional hyperparameter \\
UCB & $\beta=2.0$ \\
MES & Candidate set size $c=1000$ \\
Sobol & No surrogate \\
\bottomrule
\end{tabularx}
\end{table}

Table~\ref{tab:p3_benchmark_setup} summarizes the experimental setup. The controller has four tunable parameters. BO operates in a normalized input space $[0,1]^4$, while the task evaluates the corresponding parameters in the original controller bounds. The observed objective values are standardized within each BO run before fitting the GP model.

All BO variants use the same surrogate model: a zero-mean GP with a Mat\'ern-$5/2$ kernel, automatic relevance determination, Gaussian likelihood, and MAP hyperparameter fitting. The only difference between the BO variants is the acquisition function. We use BoTorch's default implementations of LogEI, UCB, and MES with the hyperparameters listed in Table~\ref{tab:p3_benchmark_setup}. Sobol sampling serves as the non-model-based baseline.

We use $d+1$ initial Sobol samples, where $d=4$ is the search-space dimension, followed by a total budget of $10d=40$ evaluations. This represents a low-budget tuning regime, which is typical for expensive controller evaluations. To emulate a black-box setting, no acquisition-function hyperparameters are tuned specifically for the cart-pole task.
The main online metric is the best observed cost after each iteration,
\begin{equation}
 \min_{\kappa \in \{1,\ldots,k\}} J_\kappa .  
\end{equation}

For statistical analysis, we use 20 seeds per method. For the online best-observed cost, we report the median and the $25\%$ and $75\%$ quantiles. For the final recommended controllers, we first apply a Friedman test across all methods. If the null hypothesis is rejected, we compare the best-performing method against the remaining methods using paired two-sided Wilcoxon signed-rank tests with Holm--Bonferroni correction.

\subsection{Results and Conclusion}

\begin{figure}[!t]
    \centering
    \includegraphics[width=0.9\linewidth]{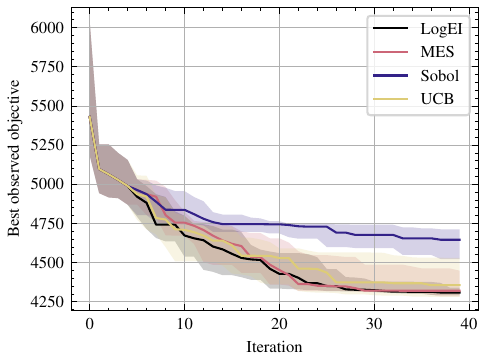}
    \captionsetup{width=\linewidth,font=footnotesize}
    \caption{Best observed cost over the optimization budget for the cart-pole benchmark. Lines show the median over seeds; shaded regions show the $25\%$ and $75\%$ quantiles.}
    \label{fig:p3convegence}
\end{figure}

Figure~\ref{fig:p3convegence} shows the behavior of the optimization algorithms. The BO variants improve faster than Sobol sampling on this task. Among the tested methods, LogEI achieves the lowest final median cost.

Table~\ref{tab:performance} summarizes the final performance. The Friedman test indicates that at least one method differs significantly from the others. In the post-hoc comparison, LogEI performs significantly better than Sobol sampling. However, the differences between LogEI and the other BO acquisition functions are not statistically significant under the chosen test. Thus, the results should not be interpreted as evidence that LogEI is generally superior to UCB or MES.

\begin{table}[!t]
\centering
\captionsetup{width=\linewidth,font=footnotesize}
\caption{Final performance of the recommended controller. The bold value indicates the best method. Underlined values are not significantly worse than the best method under the chosen post-hoc test. The runtime row reports the average wall-clock time for one complete optimization run including simulation time.}
\label{tab:performance}
\scriptsize
\renewcommand{\arraystretch}{1.12}
\setlength{\tabcolsep}{2.5pt}
\begin{tabularx}{\columnwidth}{@{}>{\raggedright\arraybackslash}p{0.23\columnwidth}*{4}{>{\centering\arraybackslash}X}@{}}
\toprule
\rowcolor{myblue!42}
Method & Sobol & LogEI & MES & UCB \\
\midrule
Med.$\pm$IQR
& $4647{\pm}190$
& $\mathbf{4310{\pm}48}$
& \underline{$4321{\pm}165$}
& \underline{$4357{\pm}250$} \\
Runtime (s)
& $\mathbf{16}$
& $31$
& $57$
& $31$ \\
\bottomrule
\end{tabularx}
\end{table}

\begin{figure}[!t]
    \centering
    \includegraphics[width=\linewidth]{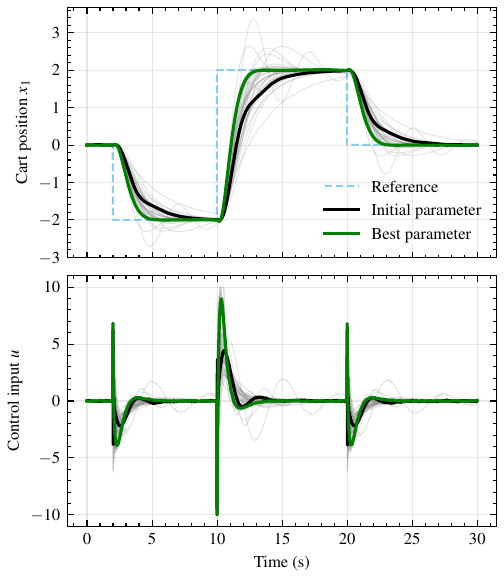}
    \captionsetup{width=\linewidth,font=footnotesize}
    \caption{State trajectories of one optimization run with the LogEI acquisition function.}
    \label{fig:p3trajectories}
\end{figure}

Finally, Figure \ref{fig:p3trajectories} shows the state trajectories of one optimization run for LogEI for the median performance. 

This minimal benchmark supports two conclusions. First, BO can outperform Sobol sampling on this cart-pole tuning task under a small evaluation budget. Second, the choice of acquisition function does not significantly influence performance in this specific case. This motivates broader benchmarking across multiple controller tuning and robotics tasks.

\section{Future Research Directions} \label{sec:p3futureissues}

Bayesian optimization has become a widely used tool for learning in control and robotics, but substantial opportunities remain for future research.

\fakepar{Benchmarking} 
Section~\ref{sec:p1otherparadigms} reviews the types of problems for which BO is particularly well suited, yet rigorous empirical comparisons with other learning and optimization paradigms remain scarce. For instance, the sample efficiency of BO should be systematically compared with that of other learning-based control approaches, such as model-based reinforcement learning. Similarly, comparisons between BO and other surrogate-based black-box optimization methods are lacking.

Section~\ref{sec:p2vanillaBO} highlights key design choices in vanilla BO. While GPs
dominate as surrogate models in the BO for control and robotics literature, other domains commonly rely on alternatives such as random forests or tree-structured Parzen estimators. Another promising alternative are neural network models such as VBLLs. More comprehensive benchmarking across a broader range of controller-tuning and robot-learning tasks is needed to determine which surrogate models are most appropriate in this setting. The same holds for GP modeling decisions, including kernel choice, mean function, likelihood, and hyperparameter treatment.

Moreover, as discussed in Sec.~\ref{p3:howtobenchmark}, realistic benchmarks for control and robotics learning remain limited. 
To this end, we provide a lightweight benchmarking framework \textsc{TuneControl} (\href{https://github.com/Data-Science-in-Mechanical-Engineering/tunecontrol}{\faGithub})
and present an example in Sec.~\ref{sec:p3examplebenchmark}. With this, we hope to encourage the community to contribute their own benchmark problems to \textsc{TuneControl}.
This shall allow our community to evaluate BO design choices for control and robotics more rigorously, and thus increase visibility and impact of BO research.

\fakepar{Methods}
The last decade has seen substantial advances in BO methods, as this paper surveys. Some of those methods have seen many successful applications ranging from lab-scale demonstrators to industry scale problems. Still, there is large potential for advancements, partly driven by open problems and partly by new opportunities from ML. We outline some examples below:

Safe BO (Sec.~\ref{sec:p2safeBO}) is a prominent BO research direction. However, most hardware applications use fixed GP-hy\-per\-pa\-rameters and specific values for $\beta$. An important step in safe BO research would be to show in a use case beyond lab-scale demonstrators how those parameters can be determined prior to optimization. Additionally, further developing a common understanding and categorizing different notions of \emph{safety}, \eg deterministic/stochastic guarantees, cautious exploration, etc., and their practical requirements would help to bridge the gap between methods development and application.

At this point BO is largely bound to the strict episodic setting mostly limiting its application to highly repetitive tasks. Contextual, time-varying and early-stopping BO provide promising mechanisms to relax this assumption, but further research is needed to transfer BO’s strong sample efficiency to settings that are less repetitive and more dynamic. 

Recent work has explored several approaches to combining BO with foundation-model paradigms, including large language models and prior-fitted networks. Although applications in control and robotics remain limited, this line of research is highly relevant to these fields, as it offers a way to integrate priors learned from large-scale data sources with the sparse, task-specific data available during optimization.

Finally, combining BO with ideas from other learning-based control methods, as well as with alternative black-box optimization paradigms, remains a promising avenue with significant potential for methodological and practical advances.

\section*{Declaration of generative AI %
in the manuscript preparation process.}

During the preparation of this work the author(s) used various LLMs to assist with code generation and correction of grammar and spelling. After using each tool/service, the author(s) reviewed and edited the content as needed and take(s) full responsibility for the content of the published article.

\bibliographystyle{cas-model2-names}

\bibliography{literature}

\end{document}